\documentclass{article}

 \usepackage[preprint]{neurips_2026}

\usepackage[utf8]{inputenc} % allow utf-8 input
\usepackage[T1]{fontenc}    % use 8-bit T1 fonts
\usepackage{hyperref}       % hyperlinks
\usepackage{url}            % simple URL typesetting
\usepackage{booktabs}       % professional-quality tables
\usepackage{amsfonts}       % blackboard math symbols
\usepackage{nicefrac}       % compact symbols for 1/2, etc.
\usepackage{microtype}      % microtypography
\usepackage{ulem}           % for \sout (strikeout text)
\usepackage{nicefrac}       % compact symbols for 1/2, etc.
\usepackage{microtype}      % microtypography
\usepackage[table]{xcolor}  % colors and table row shading
\usepackage{amsmath}        % math environments
\usepackage{amssymb}        % math symbols
\usepackage{graphicx}
\usepackage{algorithm}
\usepackage{subcaption}
\usepackage{algorithmic}
\usepackage{float}
\usepackage{wrapfig}
\usepackage{booktabs}
\usepackage{multirow}
\usepackage{array}

\usepackage{xcolor}

\definecolor{DarkGreen}{RGB}{0,120,0}
\definecolor{DarkRed}{RGB}{180,40,40}

\newcommand{\imp}[1]{{\color{DarkGreen}(#1)}}
\newcommand{\worse}[1]{{\color{DarkRed}(#1)}}

\title{Force-Aware Neural Tangent Kernels for Scalable and Robust Active Learning of MLIPs}

 \author{
Eszter Varga-Umbrich$^{1, 2}$,
Zachary Weller-Davies$^{1, \dagger}$,
Paul Duckworth$^{1}$ \\[0.5em]
\textbf{Jules Tilly$^{1}$,
Olivier Peltre$^{1}$,
Shikha Surana$^{1, \dagger, *}$} \\
\small $^{1}$InstaDeep, UK \\
\small $^{2}$Department of Engineering, University of Cambridge, Cambridge, UK \\
\small $^{\dagger}$These authors supervised equally. \\
\small $^{*}$Corresponding author: s.surana@instadeep.com
}

\begin{document}

\maketitle
\begin{abstract}
Active learning for machine-learning interatomic potentials (MLIPs) must address several challenges to be practical: scaling to large candidate pools, leveraging energy–force supervision, and maintaining robustness when candidate pools are biased relative to the target distribution. In this work, we jointly address these challenges. We first introduce a linearly scaling acquisition framework based on chunked feature-space posterior-variance shortlisting. By avoiding materialisation of the candidate and train set kernels, this approach enables screening of \(\sim\)200k structures within hours and applies broadly to acquisition strategies that score candidates based on molecular similarity metrics. We then extend the Neural Tangent Kernel (NTK) to a force-aware setting via mixed parameter–coordinate derivatives, yielding a force NTK and a joint energy–force NTK that provide natural similarity metrics for vector-field prediction. We demonstrate the effectiveness of the joint energy–force NTK on the OC20 dataset, where force-aware acquisition is crucial: it achieves the lowest energy and force MAE and RMSE across all metrics and distribution splits. Across T1x, PMechDB, and RGD benchmarks, our force NTK methods remain competitive with established baselines while being significantly more efficient than committee-based approaches. Under a controlled candidate-pool shift case study on T1x, acquisition based on pretrained MLIP embeddings and NTKs remains robust, whereas committee-based methods exhibit higher variance. Overall, these results show that a single pretrained MLIP can enable scalable, force-aware, and distribution-robust active learning for foundation-model fine-tuning.
\end{abstract}

\section{Introduction}

Machine-learning interatomic potentials (MLIPs)~\cite{behler2007generalized} predict energies and forces at orders-of-magnitude lower cost than density functional theory (DFT) and are now standard tools for atomistic simulation across molecular and materials systems~\cite{Jacobs_2025,WANG2024109673,jmi.2025.17}. Their accuracy is tightly coupled to the coverage and quality of the training data, a dependence that persists for pretrained atomistic foundation models~\cite{wood2026umafamilyuniversalmodels,batatia2023macemp}: downstream accuracy still hinges on whether the curated fine-tuning set covers the target task~\cite{anstine2025data}. Modern simulation workflows, including molecular dynamics, metadynamics~\cite{Laio2002}, nudged elastic band methods~\cite{Jonsson1998,Henkelman2000}, and uncertainty-driven samplers~\cite{zaverkin2024uncertaintybiased}, routinely produce thousands to millions of candidate structures, since proposing geometries is orders of magnitude cheaper than DFT labelling. Dataset curation is therefore the practical bottleneck for deploying MLIPs in new chemical domains. Active learning (AL)~\cite{Settles2012} addresses this bottleneck by prioritising informative candidates for labelling; in offline pool-based AL the unlabelled pool is fixed in advance~\cite{zaverkin2022batch_conformations,DDP2026}, decoupling candidate generation from acquisition.

Effective offline pool-based AL for MLIPs raises three coupled challenges. First, acquisition must scale to candidate pools containing hundreds of thousands of structures: committee ensembles add a retraining cost that grows with the labelled set, while kernel-based methods become memory-bound once similarity matrices no longer fit in memory. Second, MLIP labels comprise both scalar energies and per-atom forces, and the two are not interchangeable acquisition signals: forces, not energies, govern molecular dynamics and geometry relaxation, and a structure that most reduces force error need not be the one with the most uncertain energy. Energy-only acquisition can therefore systematically miss force-informative candidates. Third, candidate pools inherit the bias of the generators that produce them: molecular dynamics oversamples near-equilibrium basins, reaction-path methods emphasise chosen endpoints, and enhanced samplers encode bias through their collective variables and initialisation, so the pool distribution can differ sharply from the deployment target~\cite{deng2025softening,cui2025tta}.

Recent work has shown that model-based kernels constructed from pretrained equivariant MLIPs~\cite{pretrained}, specifically energy NTKs~\cite{jacot2018ntk,zaverkin2022batch_conformations} and hidden-activation features~\cite{Ouyang2024}, outperform committee and fixed-descriptor baselines on small reactive-chemistry pools. Two limitations, however, constrain their applicability in large scale and diverse settings: pairwise similarity evaluations scale as \(O(n_{\mathcal{T}} n_{\mathcal{P}})\) and \(O(n_{\mathcal{P}}^2)\) in the training-set $(n_{\mathcal{T}})$ and candidate-pool $(n_{\mathcal{P}})$ sizes, and existing constructions capture only energy sensitivity. Neither strategy has been evaluated under controlled candidate-pool bias.

We make the following contributions that together enable model-based acquisition to operate effectively in large, biased candidate pools:
\begin{itemize}
    \item \textbf{Scalable feature-space acquisition via shortlisting:}
    We introduce a feature-space posterior-variance (PV) pipeline with shortlist-based selection that remains conditioned on the training set while avoiding \(O(n_{\mathcal{T}} n_{\mathcal{P}})\) train--pool and \(O(n_{\mathcal{P}}^2)\) pool--pool kernel computations. Candidates are scored in chunks using explicit \(d\)-dimensional feature representations, and a top-\(K\) subset is first selected by PV acquisition in \(O(n_{\mathcal{T}} + n_{\mathcal{P}})\) time. A diversity-based Largest Cluster Maximum Distance (LCMD) rule~\cite{holzmuller2023batch_al} is then applied to this shortlist to select a batch of size \(B\) in \(O(K n_{\mathcal{T}})\). By shortlisting in feature space, this approach scales linearly in both the training and pool sizes, enabling acquisition on \(\sim\!200\)k candidates and, in principle, much larger pools.

    \item \textbf{Force-aware NTK representations:}
    We extend the Neural Tangent Kernel~\cite{jacot2018ntk, zaverkin2022batch_conformations, pretrained} to force prediction by incorporating mixed parameter--coordinate derivatives \(\nabla_\theta \nabla_{\mathbf{r}} E_\theta\), yielding a force-aware NTK and a joint energy--force NTK. The joint representation subsumes the energy and force NTKs as special cases and enables interpolation between energy- and force-driven acquisition.

    \item \textbf{Robustness and scaling analysis:}
    We provide a systematic evaluation across reactive molecular benchmarks (T1x~\cite{Schreiner2022}, PMechDB~\cite{Tavakoli2024}, and RGD~\cite{Zhao2023}) and large-scale catalysis (OC20~\cite{Chanussot_2021,sahoo2025opencatalyst2025oc25}), including controlled candidate-pool bias experiments on T1x that isolate inter- and intra-reaction shifts. We show that model-derived kernels remain robust under pool shift, while committee-based methods exhibit higher variance due to their reliance on point-wise uncertainty. We additionally characterise the practical operating regime of our methods with respect to pool size and feature dimension.
\end{itemize}

On OC20, we find that incorporating force information is essential for effective active learning. The joint energy–force NTK combined with the shortlist-LCMD acquisition achieves the lowest energy and force errors across all metrics, outperforming competing methods on both in- and out-of-distribution splits. However, we find that force-awareness is not always essential: across the reactive benchmarks, energy-based acquisition already captures most of the informative variation. Incorporating force information does not provide additional benefit; however, the force-only and joint energy–force NTKs still track the performance of energy-only NTKs closely, with the energy-only variant remaining the strongest baseline in this setting. Taken together, these results highlight the importance of methods that generalise across both regimes and scales. Overall, we find that the joint energy–force NTK provides a consistently strong and robust acquisition strategy across the benchmarks considered in this work.

\section{Related Work}

\paragraph{Active learning for MLIPs:}
Most MLIP active-learning workflows are online, where structures are generated during molecular dynamics, reaction-path exploration, or perturbations and selected using uncertainty or extrapolation criteria from the current model. Representative approaches include committee-based methods~\cite{smith2018less,schran2020committee,Niblett2025}, extrapolation criteria for moment-tensor potentials~\cite{podryabinkin2017active}, Bayesian and Gaussian-process force fields~\cite{jinnouchi2019mlff,vandermause2020flare}, and concurrent or uncertainty-driven sampling schemes~\cite{zhang2019uniformly,kulichenko2023uncertainty,zaverkin2024uncertaintybiased,zaverkin2026metadynamics}. Offline pool-based active learning instead selects from a fixed candidate set~\cite{zaverkin2022batch_conformations,DDP2026} and is increasingly relevant for dataset curation and distillation from large atomistic datasets~\cite{levine2026openmolecules2025omol25,sahoo2025opencatalyst2025oc25,Horton2025}. In this setting, candidate pools can be large and biased relative to the target distribution, which can degrade acquisition performance under distribution shift~\cite{deng2025softening,cui2025tta}.

\paragraph{Acquisition signals and model-based kernels:}
Common acquisition strategies include committee disagreement~\cite{schran2020committee,peterson2017uncertainty}, as well as more efficient variants such as shallow ensembles~\cite{kellner2024uq_shallow}, multi-head pretrained models~\cite{beck2025multihead}, learned uncertainty heads~\cite{neumann2025orbv3} and Bayesian interatomic potentials \cite{jinnouchi2019mlff,vandermause2020flare, coscia2026blipsbayesianlearnedinteratomic}. Descriptor-based methods that use fixed representations such as SOAP~\cite{bartok2013representing,de2017mlunifies,dscribe} or molecular fingerprints with Tanimoto similarity~\cite{Morgan1965,rogers2010extended,Ralaivola2005} are typically combined with diversity objectives~\cite{DDP2026,kellner2025enhanced_sampling}. More recently, model-based kernels derived from neural representations, such as energy NTK features~\cite{zaverkin2022batch_conformations} and activation kernels~\cite{pretrained, Ouyang2024}, have been shown to be useful signals for pretrained equivariant MLIPs (see \cite{holzmuller2023batch_al} for a general discussion on deep batch AL). However, existing demonstrations that incorporate diversity still compute full train--pool or pool--pool similarity computations~\cite{zaverkin2022batch_conformations, holzmuller2023batch_al}. Feature-space posterior-variance methods are well known~\cite{Rasmussen2005, holzmuller2023batch_al}, and have enabled AL for candidate pools of up to \(\sim\!150\)k \cite{zaverkin2022batch_conformations}, but utilising them for efficient shortlisting has not been explored in the context of MLIPs.

%While random-projection approximations have enabled candidate pools of up to \(\sim\!150\)k structures~\cite{zaverkin2022batch_conformations}, these approaches still rely on explicit pairwise train-pool similarity evaluations. Feature-space posterior-variance methods are well known~\cite{Rasmussen2005, holzmuller2023batch_al}, and have been used but have not previously been explored for scalable active learning of MLIPs via shortlisting.

%However, these approaches have so far only considered energy-aware acquisition, and have been limited to relatively small candidate pools (\(\sim\!10\)k structures), as they rely on materialising the full candidate--candidate kernel, which scales as \(O(n_{\mathcal{P}}^2)\). %Our work removes this limitation by implementing an acquisition strategy that scales linearly in the pool size, enabling model-based acquisition to scale to large candidate pools while extending these representations to force-aware settings.

\section{Methods}

\subsection{Active learning}

We study offline pool-based active learning for MLIPs. The learner is given a small labelled training set $\mathcal{T}^{(0)}$ and a large fixed pool of unlabelled candidate structures $\mathcal{P}^{(0)}$, together with a per-round labelling budget $B$ and a total budget of $T$ rounds. 
Each candidate is an atomic structure $\mathbf{x} = (\mathbf{z}, \mathbf{r})$, with atomic numbers $\mathbf{z} = (z_i)_{i=1}^{N(\mathbf{x})}$ and Cartesian coordinates $\mathbf{r} = (\mathbf{r}_i)_{i=1}^{N(\mathbf{x})}$, where $N(\mathbf{x})$ is the number of atoms. The label $y(\mathbf{x})$ associated with a structure consists of its total energy $E(\mathbf{x}) \in \mathbb{R}$ and per-atom forces $\mathbf{F}_i(\mathbf{x}) \in \mathbb{R}^3$, related by $\mathbf{F}_i(\mathbf{x}) = -\nabla_{\mathbf{r}_i} E(\mathbf{x})$, and is obtained from a reference DFT calculation.

At each round $t = 0, 1, \dots, T-1$, an acquisition function $\alpha$ scores candidates in $\mathcal{P}^{(t)}$ and selects a batch
\begin{equation}
\mathcal{A}^{(t)} = \operatorname*{arg\,max}_{\mathcal{A} \subseteq \mathcal{P}^{(t)},\, |\mathcal{A}| = B} \alpha(\mathcal{A}),
\end{equation}
where $\alpha$ may depend on the current model $f_{\theta^{(t)}}$, the labelled set $\mathcal{T}^{(t)}$, or fixed structural descriptors of the candidates, depending on the acquisition method. Reference labels are then queried for the selected candidates, and the training set and pool are updated as

\begin{equation}
\mathcal{T}^{(t+1)} = \mathcal{T}^{(t)} \cup \{(\mathbf{x}, y(\mathbf{x})) : \mathbf{x} \in \mathcal{A}^{(t)}\}, \qquad \mathcal{P}^{(t+1)} = \mathcal{P}^{(t)} \setminus \mathcal{A}^{(t)}.
\end{equation}

The model is fine-tuned on $\mathcal{T}^{(t+1)}$ before the next round. The goal is to choose acquisition batches that minimise test error after $T$ rounds, given the fixed budget $T \cdot B$.

Our experiments use pretrained MACE models~\cite{batatia2023macehigherorderequivariant} trained via the \texttt{mlip} library~\cite{brunken2025machinelearninginteratomicpotentials} via \texttt{jax}~\cite{jax2018github}. The MACE architecture predicts total energies from input structures $E_{\theta}(\mathbf{x})$, and forces are obtained by auto-differentiation. 

\subsection{Acquisition signals}\label{sec:acquisition}

Each candidate label consists of a scalar energy and atomic forces, so an effective acquisition signal should reflect sensitivity to both quantities. The methods we compare fall into three families: model-based kernels derived from the MLIP (energy NTK, force NTK, energy--force NTK, and latent activations), descriptor-based kernels using fixed structural representations (SOAP~\cite{bartok2013representing} and Morgan fingerprints~\cite{Morgan1965,rogers2010extended}), and direct uncertainty scores from committee disagreement, alongside random selection. Kernel-based methods construct a similarity geometry over the candidate pool and pair it with a batch selection rule (Section~\ref{sec:selection}); direct methods assign point-wise scores and do not encode redundancy or coverage among candidates.

Kernel methods are derived from an underlying molecular representation \(\phi(\mathbf{x}) \in \mathbb{R}^{d}\). For binary representations such as Morgan fingerprints, we use Tanimoto similarity~\cite{rogers2010extended}. For all other embeddings, we define kernels via inner products \(k(\mathbf{x}, \mathbf{x}') = \phi(\mathbf{x})^\top \phi(\mathbf{x}')\) of cosine-normalised embeddings.

\paragraph{Energy NTK (NTK-E):}
Following prior work~\cite{zaverkin2022batch_conformations, pretrained}, we represent a structure by the sensitivity of the predicted energy to model parameters,
\begin{equation}\label{eq: energy_ntk}
\phi_E(\mathbf{x}) = \nabla_{\theta} E_\theta(\mathbf{x}).
\end{equation}
Computing $\nabla_{\theta} E_\theta(\mathbf{x})$ for a single structure requires only a standard backward pass, but the resulting feature has dimension equal to the full parameter count, so storing the $d \times d$ feature-space precision matrix or materialising pairwise NTK entries quickly becomes memory-prohibitive. We therefore restrict the NTK to a parameter subspace $\theta_P$, with the embedding parameter blocks of MACE used as the default choice (see Appendix~\ref{app:mace-architecture}). Although these weights are species-indexed, $\nabla_{\theta_P} E_\theta(\mathbf{x})$ is computed by backpropagating through every subsequent interaction and readout layer, so the NTK aggregates whole-model sensitivities into a feature that distinguishes structures by both composition and geometry. Parameter subsets trade-offs are compared in Section~\ref{sec:scaling}.

\paragraph{Force NTK (NTK-F):}
NTK-E does not directly capture sensitivity to forces. To extend the NTK to vector-field prediction, we consider the mixed parameter-coordinate derivative
\begin{equation}\label{eq: forcentk}
\mathbf{J}_{\theta_P}(\mathbf{x}) = \nabla_{\theta_P} \mathbf{F}(\mathbf{x})
= -\nabla_{\theta_P}\nabla_{\mathbf{r}} E_\theta(\mathbf{x}),
\end{equation}

Since $\mathbf{J}_{\theta_P}$ depends on the atomic positions through atomic forces, to get a structure level embedding, we must aggregate over the atoms. For any translationally invariant force field, $\sum_i \textbf{F}_i=0$, and so we instead aggregate via the root mean square 
\begin{equation}
    \phi_{F}(\mathbf{x}) = \frac{1}{\sqrt{N(\mathbf{x})}}\sqrt{ \mathbf{J}_{\theta_P}(\mathbf{x}) \cdot \mathbf{J}_{\theta_P}(\mathbf{x}) },
\end{equation}
where the dot product is taken over coordinate space and not parameter space. The resulting embedding $\phi_{F}(\mathbf{x})$ is thus an equivariant molecular embedding with dimension equal to the number of parameters $\theta_P$.

\paragraph{Energy-force NTK (NTK-EF):}
Energy and force sensitivity need not rank candidates identically (see Appendix~\ref{app:force_ntk_plot} for a visual comparison of energy and force NTK kernels), and a general molecular embedding should be explicitly sensitive to both energy and force prediction. We therefore combine them into a single representation,
\begin{equation}\label{eq: efntk}
\phi_{EF}(\mathbf{x}) =
\left[\sqrt{w_E}\,\phi_E(\mathbf{x}), \sqrt{w_F}\, \phi_F(\mathbf{x})\right],
\end{equation}
where each component is normalized separately and \(w_E, w_F \ge 0\) control the relative contributions of energy and force signals. The corresponding kernel takes the form 
\begin{equation}\label{eq: kernel_full}
    K_{EF}(\mathbf{x}, \mathbf{x}') =\phi_{EF}(\mathbf{x})^\top \phi_{EF}(\mathbf{x}') = w_E k_E(\mathbf{x}, \mathbf{x}') + w_F k_F(\mathbf{x}, \mathbf{x}'),
\end{equation}
where \(k_E(\mathbf{x}, \mathbf{x}') = \phi_E(\mathbf{x})^\top \phi_E(\mathbf{x}')\) and \(k_F(\mathbf{x}, \mathbf{x}') = \phi_F(\mathbf{x})^\top \phi_F(\mathbf{x}')\). 

As a consequence, kernel-based acquisition in Equation~\eqref{eq: efntk} subsumes both energy and force NTKs as special cases (\(w_F=0\) or \(w_E=0\)) and interpolates between them depending on which signal is most informative. Although the combined kernel is a linear sum, the contributions become coupled through conditioning on the training set: posterior-variance acquisition (Equation~\eqref{eq: gpequation}) reflects how both signals interact given the observed data. 

\paragraph{Baselines:}
We include activation features~\cite{pretrained} as a learned representation baseline. As model-independent methods, we use SOAP descriptors~\cite{bartok2013representing,de2017mlunifies} computed with \texttt{DScribe}~\cite{dscribe} (\(r_{\mathrm{cut}}=6.0\)~\AA, \(n_{\max}=8\), \(l_{\max}=6\), with outer averaging), and Morgan fingerprints~\cite{Morgan1965,rogers2010extended} (radius \(r=3\), 2048 bits) with the Tanimoto kernel~\cite{Ralaivola2005} computed using \texttt{RDKit} \cite{rdkit_2025_09_1}. For direct uncertainty methods, we use pretrained committees with \(M=3\) models initialised from the same checkpoint, with diversity introduced via independent data-order shuffling \cite{pretrained}. We report energy disagreement (Committee-E) and force disagreement (Committee-F), obtained by aggregating per-atom force variance into a structure-level score. Committee variants are evaluated using independently fine-tuned models initialised from the same pretrained checkpoint. This requires training a total of \(M+1\) models. We note that potentially improved committee variants for pretrained atomistic foundation models, such as multi-head committees~\cite{beck2025multihead}, are not considered in this work. We additionally include random selection as a baseline.

\subsection{Scalable batch selection via shortlisting}\label{sec:selection}
Given a kernel, a batch selection rule chooses $B$ candidates per round. We use Gaussian posterior variance (PV) to shortlist candidates based on uncertainty, and Largest-Cluster Maximum Distance (LCMD)~\cite{holzmuller2023batch_al} to enforce diversity within the shortlist. To make this tractable for large pools, we compute PV in the feature space, then apply LCMD to a high-uncertainty shortlist. 

\paragraph{Posterior variance.} Given a kernel $k$ and current training set $\mathcal{T}^{(t)}$, a candidate $\mathbf{x}$ has Gaussian posterior variance
\begin{equation}\label{eq: gpequation}
\sigma_t^2(\mathbf{x})
=
k(\mathbf{x},\mathbf{x})
-
k_{\mathcal{T}^{(t)}}(\mathbf{x})^\top
(K_{\mathcal{T}^{(t)}\mathcal{T}^{(t)}}+\lambda I)^{-1}
k_{\mathcal{T}^{(t)}}(\mathbf{x}),
\end{equation}
where $k_{\mathcal{T}^{(t)}}(\mathbf{x})$ is the vector of similarities between $\mathbf{x}$ and the training set. Direct evaluation requires the $n_{\mathcal{P}}^2$ candidate-candidate kernel, which is infeasible at $n_{\mathcal{P}} \sim 200$k.

\paragraph{Feature-space form.} For kernels with explicit feature maps $k(\mathbf{x}, \mathbf{x}') = \phi(\mathbf{x})^\top \phi(\mathbf{x}')$, posterior variance can be computed directly in feature space \cite{holzmuller2023batch_al}. Let $\Phi_{\mathcal{T}} \in \mathbb{R}^{n_{\mathcal{T}} \times d}$ denote the feature matrix of the labelled set, and define the feature-space precision matrix
\begin{equation}\label{eq: precision}
M_{\mathcal{T}} = (\Phi_{\mathcal{T}}^\top \Phi_{\mathcal{T}} + \lambda I)^{-1} \in \mathbb{R}^{d \times d},
\end{equation}
By the Woodbury identity~\cite{Golub2012, Rasmussen2005}, the score
\begin{equation}\label{eq: score_gp}
s(\mathbf{x}) =
\phi(\mathbf{x})^\top M_{\mathcal{T}} \phi(\mathbf{x}),
\end{equation}
is proportional to the Gaussian posterior variance and yields the same candidate ranking, while avoiding construction of the $n_{\mathcal{P}}^2$ candidate--candidate kernel.

We compute $M_{\mathcal{T}}$ by accumulating $\Phi_{\mathcal{T}}^\top \Phi_{\mathcal{T}}$ over chunks of training features, then inverting the resulting $d \times d$ matrix. 
Candidate scores $s(\mathbf{x})$ are computed in chunks against the cached $M_{\mathcal{T}}$, while a running top-$K$ shortlist tracks the highest-scoring candidates. This keeps peak memory at $O(d^2 + Cd)$ for chunk size $C$ and makes the total acquisition cost linear in $n_{\mathcal{P}}$ and $n_{\mathcal{T}}$.

% To compute $M_{\mathcal{T}}$ efficiently, we accumulate the Gram matrix \(\Phi_{\mathcal{T}}^\top \Phi_{\mathcal{T}}\) in chunks. Specifically, for chunks \(\Phi_C \in \mathbb{R}^{C \times d}\) of the training features, we maintain a running sum $\sum_C \Phi_C^\top \Phi_C$ after which the feature precision matrix can be formed. This requires materialising matrices of size at most \(O(\max(Cd, d^2))\), making the computation linear in \(n_{\mathcal{T}}\) and memory-efficient in the number of training points. Candidate scores \(s(\mathbf{x})\) for \(\mathbf{x} \in \mathcal{P}\) are also computed in chunks using the same strategy. During the scoring pass, we maintain a running top-\(K\) representation shortlist by caching the highest-scoring candidates, enabling efficient selection without re-evaluating the full pool.

\paragraph{LCMD on the shortlist.}
Posterior variance alone tends to favour isolated high-uncertainty points. We therefore use PV only to shortlist the $K$ highest-uncertainty candidates, then apply LCMD on the shortlist to select the final batch. LCMD modifies greedy farthest-point sampling to favour dense but under-covered regions, mitigating redundancy and discouraging outlier selection. This separation is deliberate: PV acts as an uncertainty-based filter with full conditioning on the training set, while LCMD enforces diversity. 

In practice, we are able to use this method to efficiently score candidate pools of $\sim 200$k points with shortlists of size $K= 50$k, but in theory, PV shortlisting extends linearly in the candidate and training pool, in principle unlocking much larger sets. Further details and pseudocode are given in Appendix~\ref{app:scalable_pv}.

\section{Results}
\label{sec:experiments}

Our experiments are designed to isolate the roles of force awareness, robustness to distribution shift, and scalability in offline active learning. We first evaluate our methods in a large-scale OC20 candidate-pool setting, where we show that force-aware acquisition becomes essential for strong performance. We then study reactive benchmarks in Section~\ref{sec:force_ntk_main} to better understand under what conditions explicit force awareness improves acquisition beyond energy-only signals. In Section~\ref{sec:cpshift}, we perform controlled candidate-pool biasing on T1x to evaluate robustness under distribution shift and study how different acquisition strategies respond to biased candidate pools. Finally, Section~\ref{sec:scaling} analyses runtime and memory costs. All experiments run on a single H100 80GB GPU.

\subsection{Large-pool OC20 catalyst benchmark}
\label{sec:oc20}

\begin{figure}[h]
\centering
\includegraphics[width=\linewidth]{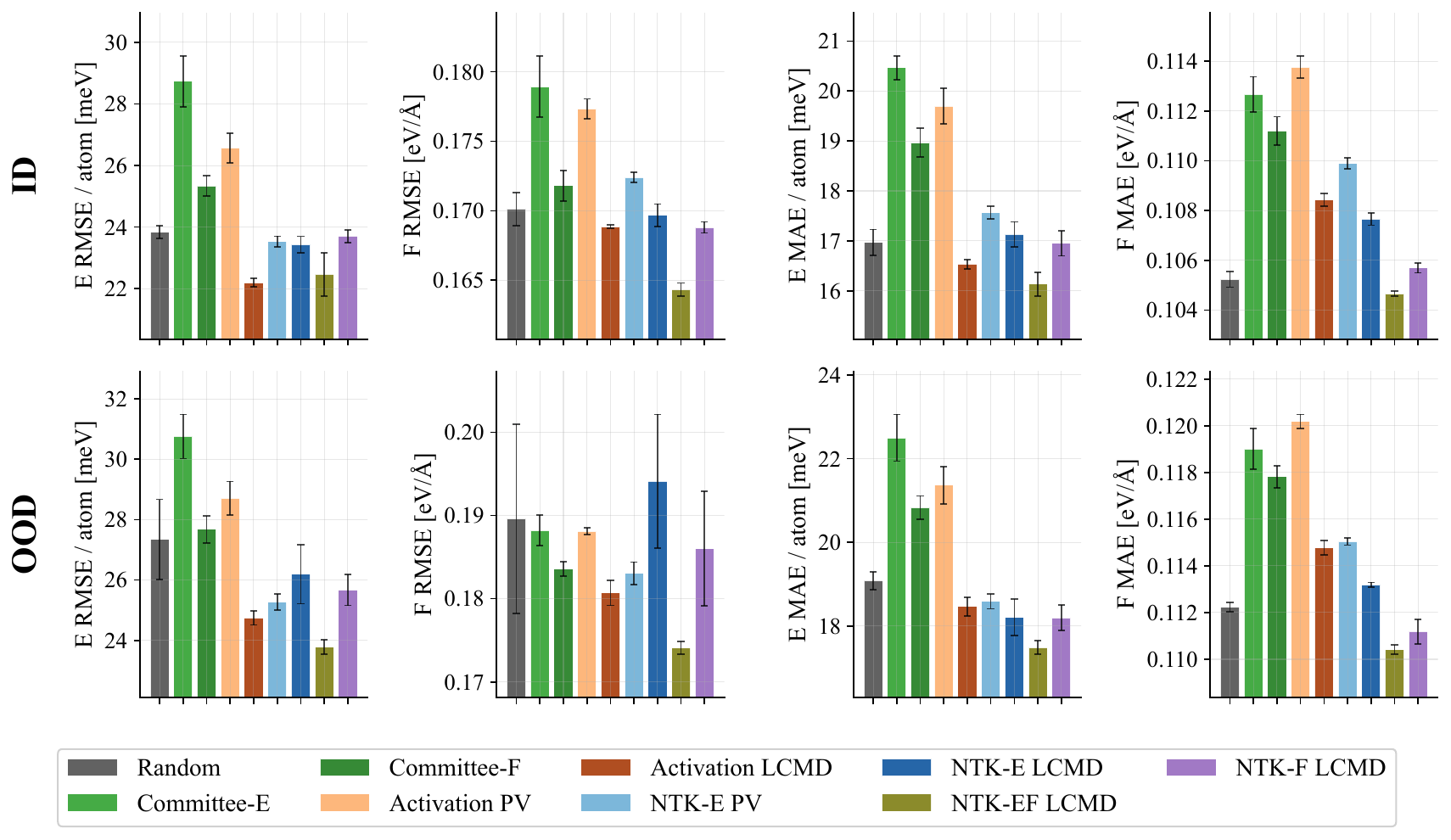}
\caption{
OC20 final-round test errors on \texttt{val\_is} (ID) and the mean of the three out-of-distribution splits \texttt{val\_oos\_ads}, \texttt{val\_oos\_bulk}, and \texttt{val\_oos\_ads\_bulk}. The joint energy--force NTK achieves the lowest energy and force errors across all panels. %The force-only NTK trails by a small but consistent margin, while the energy-only NTK remains competitive on the ID split but degrades substantially under distribution shift. Both committee variants are the weakest non-random baselines, with activation-based acquisition exhibiting similar degradation on the OOD splits.
}
\label{fig:oc20_finalbars}
\end{figure}
We consider active learning on a randomly selected $200$k subset of OC20 \cite{Chanussot_2021}. %Two features distinguish this setting from the reactive benchmarks of Section~\ref{sec:force_ntk_main}. First, the pool is almost two orders of magnitude larger, making memory-efficient acquisition essential. 
To handle the pool size we use feature-space PV to shortlist candidates, followed by LCMD selection (Section~\ref{sec:acquisition}). OC20 candidate structures are drawn from relaxation trajectories near, but not at, local minima, where small geometric perturbations drive large force changes that energy similarity alone cannot resolve.

We use a MACE model pretrained on MPTraj~\cite{deng2023chgnet}. Each run starts from \(2{,}000\) labelled structures and acquires \(1{,}250\) per round for six rounds, averaged over three seeds. Performance is evaluated on a 20k-structure test set partitioned evenly across the four OC20 regimes~\cite{Chanussot_2021}: in-distribution (\texttt{val\_is}) and three out-of-distribution splits with held-out adsorbates, bulks, or both. We use a balanced weighting \(w_E{:}w_F = 1{:}1\) for the NTK-EF.

Figure~\ref{fig:oc20_finalbars} reports final-round test errors. The NTK-EF variant attains the lowest error on every metric and every split. NTK-F trails NTK-EF by a small but consistent margin. Forces are per-atom quantities sensitive to local geometry; energies are extensive and reflect the full structure. The joint kernel inherits both views, so candidates that match in local force pattern but differ in overall composition or size remain distinguishable. Activation-LCMD is competitive on \texttt{val\_is} but degrades on the OOD splits. NTK-E degrades most sharply on \texttt{val\_oos\_ads}, where energy features alone are insufficient to distinguish adsorbate chemistry and the diversity objective ends up emphasising bulk-surface geometry rather than adsorbate variation. Its OOD energy MAE remains comparable to the other methods, while its OOD energy RMSE is the worst among non-random baselines; this gap indicates a small number of large per-structure errors rather than a broad shift in accuracy, consistent with energy-only PV occasionally selecting a few high-leverage outliers on the adsorbate-shift split. Both committee variants are the weakest non-random methods on every panel, indicating that pretrained-backbone ensemble disagreement, in either form, is not a competitive acquisition signal at this scale. This is consistent with prior reports that ensemble disagreement underestimates uncertainty under distribution shift~\cite{Kahle_2022}. With the LCMD step removed, PV acquisition performs at or near the random baseline on every split (Figure~\ref{fig:oc20_finalbars}), indicating that diversity-aware selection is required for the PV signal to translate into informative batches at this scale. Per-split breakdowns and full learning curves are in Appendix~\ref{app:oc20}.

\subsection{When does force-awareness matter?}
\label{sec:force_ntk_main}

We now test whether force sensitivity improves acquisition across reactive benchmarks. Following the protocol of \citet{pretrained}, we evaluate on three reactivity datasets: HCNO-filtered PMechDB~\cite{Tavakoli2024}, RGD~\cite{Zhao2023}, and T1x~\cite{Schreiner2022}, using a SPICE-2-pretrained~\cite{SPICE,levine2026openmolecules2025omol25} MACE model. The PMechDB and RGD candidate pools are 10k and 5k random subsets of OMol~\cite{levine2026openmolecules2025omol25}, and the T1x pool comprises 100 randomly selected reaction pathways. Each run starts from 50 labelled structures and acquires 150 per round for 20 rounds, the remaining structures are split 20\%/10\% test/validation. Results are averaged over two random seeds.

\begin{table}[h] \centering \scriptsize \setlength{\tabcolsep}{2pt} \caption{ Energy and force acquisition performance across reactive benchmarks. Reported metrics are energy/force RMSE area under the acquisition curve (AUC) and final-round RMSE. Lower is better; best values in each row are bolded. Percent differences are relative to Random. } \label{tab:reactive_summary} \begin{tabular}{llcccccccc} \toprule \textbf{Dataset} & \textbf{Metric} & \textbf{Random} & \textbf{Activation} & \textbf{Committee-E} & \textbf{Committee-F} & \textbf{NTK-E} & \textbf{NTK-F} & \textbf{NTK-EF} \\ \midrule \multirow{4}{*}{\textbf{RGD}} & E AUC & 348.0 & \textbf{326.0} \imp{-6.3\%} & 339.0 \imp{-2.6\%} & 334.0 \imp{-4.0\%} & 339.0 \imp{-2.6\%} & 338.0 \imp{-2.9\%} & 344.5 \imp{-1.0\%} \\ & F AUC & 4763.5 & 4677.0 \imp{-1.8\%} & 4659.0 \imp{-2.2\%} & \textbf{4587.5} \imp{-3.7\%} & 4651.5 \imp{-2.4\%} & 4665.5 \imp{-2.1\%} & 4684.0 \imp{-1.7\%} \\ & Final E & 13.0 & 12.0 \imp{-7.7\%} & 12.0 \imp{-7.7\%} & \textbf{11.5} \imp{-11.5\%} & 12.0 \imp{-7.7\%} & 12.0 \imp{-7.7\%} & 12.0 \imp{-7.7\%} \\ & Final F & 188.5 & 185.0 \imp{-1.9\%} & 186.0 \imp{-1.3\%} & \textbf{182.5} \imp{-3.2\%} & 186.0 \imp{-1.3\%} & 184.0 \imp{-2.4\%} & 186.0 \imp{-1.3\%} \\ \midrule \multirow{4}{*}{\textbf{T1X}} & E AUC & 246.5 & 164.0 \imp{-33.5\%} & 268.5 \worse{+8.9\%} & 250.0 \worse{+1.4\%} & \textbf{159.5} \imp{-35.3\%} & 178.5 \imp{-27.6\%} & 175.0 \imp{-29.0\%} \\ & F AUC & 2300.0 & 1789.0 \imp{-22.2\%} & 2516.0 \worse{+9.4\%} & 2182.5 \imp{-5.1\%} & \textbf{1754.5} \imp{-23.7\%} & 1802.5 \imp{-21.6\%} & 1796.5 \imp{-21.9\%} \\ & Final E & 6.5 & \textbf{5.0} \imp{-23.1\%} & 5.5 \imp{-15.4\%} & \textbf{5.0} \imp{-23.1\%} & \textbf{5.0} \imp{-23.1\%} & 5.5 \imp{-15.4\%} & \textbf{5.0} \imp{-23.1\%} \\ & Final F & 62.0 & \textbf{57.0} \imp{-8.1\%} & 60.0 \imp{-3.2\%} & \textbf{57.0} \imp{-8.1\%} & \textbf{57.0} \imp{-8.1\%} & 59.5 \imp{-4.0\%} & 60.5 \imp{-2.4\%} \\ \midrule \multirow{4}{*}{\textbf{PMechDB}} & E AUC & 453.0 & 427.0 \imp{-5.7\%} & 493.0 \worse{+8.8\%} & 457.5 \worse{+1.0\%} & \textbf{426.5} \imp{-5.9\%} & 433.5 \imp{-4.3\%} & 431.0 \imp{-4.9\%} \\ & F AUC & 2857.0 & 2800.5 \imp{-2.0\%} & 2799.0 \imp{-2.0\%} & \textbf{2634.5} \imp{-7.8\%} & 2718.5 \imp{-4.9\%} & 2780.5 \imp{-2.7\%} & 2777.0 \imp{-2.8\%} \\ & Final E & 17.0 & \textbf{16.0} \imp{-5.9\%} & 19.0 \worse{+11.8\%} & 17.0 (0.0\%) & \textbf{16.0} \imp{-5.9\%} & \textbf{16.0} \imp{-5.9\%} & 16.5 \imp{-2.9\%} \\ & Final F & 113.5 & 109.0 \imp{-4.0\%} & 107.0 \imp{-5.7\%} & \textbf{102.0} \imp{-10.1\%} & 106.5 \imp{-6.2\%} & 111.5 \imp{-1.8\%} & 109.5 \imp{-3.5\%} \\ \bottomrule \end{tabular} \end{table}

% Following \cite{pretrained}, we consider three reactivity datasets: HCNO-filtered PMechDB~\cite{Tavakoli2024}, RGD~\cite{Zhao2023}, and T1x~\cite{Schreiner2022}. For RGD and PMechDB, we use random 5k and 10k subsets drawn from OMol~\cite{levine2026openmolecules2025omol25}. The T1x subset consists of 100 randomly selected reaction pathways. In this section, we use SPICE-2-pretrained~\cite{SPICE, levine2026openmolecules2025omol25} MACE models, with all labels computed at the \(\omega\)B97M-V/def2-TZVPD level of theory. We perform 20 acquisition rounds, selecting 150 points per round, starting from an initial seed set of 50 structures. For each dataset, we use 20\%/10\% test/validation splits, with the remaining structures forming the candidate pool. Results are averaged over two random seeds.

Table~\ref{tab:reactive_summary} summarises the energy and force RMSE area under the (learning) curve (AUC), together with the final-round energy and force RMSEs across the three datasets for a range of baselines (see Appendix \ref{app:reactivity} for additional comparisons against fixed-descriptor methods). All four model-based kernels (activation, NTK-E, NTK-F, NTK-EF) substantially outperform random selection, but differences between them are dataset-dependent and minute on most metrics (see Appendix \ref{app:reactivity}). Among these methods, activation features and energy NTKs provide the strongest and most consistent baselines overall. Committee-based methods, especially force committees, can perform extremely well (best in class) in some settings, but exhibit substantially higher variance across datasets and across the energy–force trade-off. Appendix \ref{app:reactivity} shows that committee MAE performance is well below random baselines, suggesting that a portion of the RMSE gains may arise from preferential selection of outlier or high-error
structures. Across model-based methods, we find that improvements over random are generally smaller for MAE than for RMSE but still positive.

Overall, these results indicate that, on the reactive benchmarks considered here, energy-based acquisition already captures much of the informative variation in the candidate pool. To support this, in Appendix \ref{app:reactivity} we find that increasing the $w_{E}: w_{F}$ ratio gives better results. %In this sense, the role of force-aware acquisition is therefore limited to refining the selection within geometrically similar regions (see Appendix \ref{app:force_ntk_plot} for a visualisation), which does not necessarily translate into improved uncertainty quantification. 
Importantly, however, the resulting degradation is very small: force-only and joint energy–force NTKs continue to closely track the performance of the energy-only NTK across datasets, while remaining substantially better than random selection. More generally, the benefit of force-aware acquisition could depend on how much additional discrimination forces provide beyond energies, which is primarily a dataset property: within a reactive pathway, energies and forces co-vary and force features add little; across heterogeneous chemistries they carry information that energy similarities cannot recover. A second factor is force-label quality: \citet{domantas} report systematic force-component errors on molecular datasets, and force acquisition signals therefore operate on a noisier label stream than the corresponding energy channel. Improved force-aware uncertainty quantification may be achievable also through alternative normalisations of the energy and force representations (see Appendix \ref{app:force_ntk_plot}); we leave this for future work.
%In Appendix \ref{app:reactivity}, we observe that increasing the relative energy weighting $w_{E}: w_{F}$ generally improves performance here, although the differences remain small, and the overall behaviour is qualitatively the same across weightings.

\subsection{Robustness to candidate pool shift}\label{sec:cpshift}

While the robustness of acquisition is implicitly evaluated by evaluating AL performance across a variety of datasets, we further study robustness to candidate pool distribution shift in a controlled setting: this is relevant since real candidate pools are biased due to the inherent biases of the structure generators. T1x \cite{Schreiner2022} provides a natural setting for studying pool shift: each reaction family contains structures along NEB and climbing-image NEB paths, so the frame index serves as a proxy position along the reaction coordinate. This allows us to simulate two forms of bias: (i) inter-reaction imbalance, which skews sampling across reactions, and (ii) intra-reaction imbalance, which biases sampling along the reaction coordinate. 
% For inter-reaction bias, we up-weight individual reactions by a factor of 5 relative to a uniform (1:1:1:1:1) baseline. For intra-reaction bias, we similarly upweight specific frames (0th, 5th, and 8th) by a factor of 5 relative to uniform sampling along the reaction coordinate. 

Across both settings, we find that representation-based acquisition methods, including both learned model-derived representations and fixed descriptors, generally exhibit more stable behaviour under candidate-pool bias than uncertainty-only approaches, particularly under intra-reaction bias along the reaction coordinate. %Activation- and NTK-based methods consistently perform well overall, with activation features yielding the lowest AUC under both inter- and intra-reaction bias, while NTK variants attain the lowest final-round errors. 
Combined with the results on OC20 and the reactive benchmarks, where committee-based methods exhibit higher variance across datasets and objectives, these findings suggest that representation-aware acquisition strategies can provide more robust behaviour under candidate-pool shift than those using only point-wise uncertainty estimates. Full experimental details and results are provided in Appendix~\ref{sec:appendix-t1x-case-study}.

\subsection{Cost, scalability, and operating range}
\label{sec:scaling}
\begin{figure}[h]
  \centering
  \includegraphics[scale=0.4]{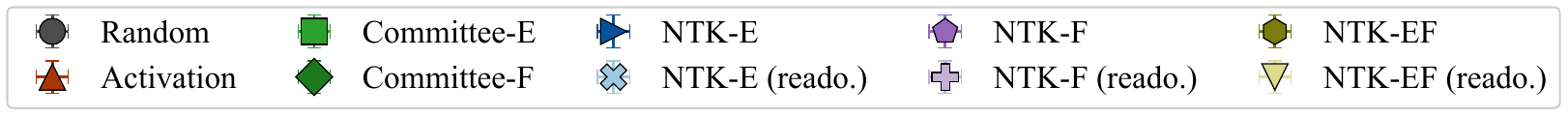}\\[2pt]
  \begin{subfigure}[t]{0.32\linewidth}
      \centering
      \includegraphics[width=\linewidth]{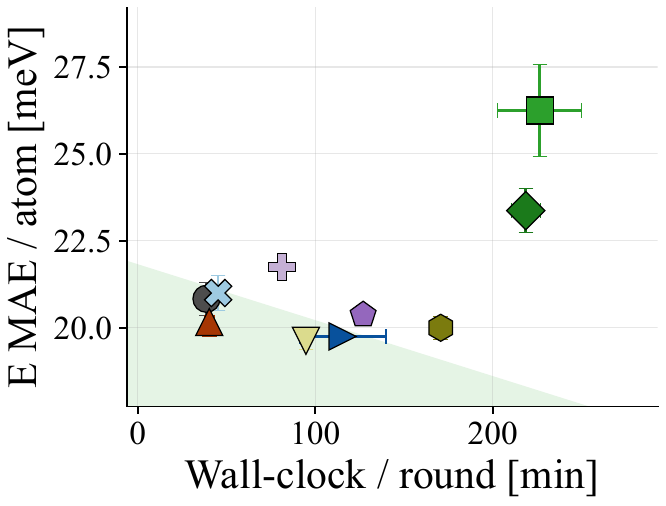}
      \caption{Energy MAE vs cost.}
      \label{fig:pareto_oc20_MAEenergy}
  \end{subfigure}
  \hfill
  \begin{subfigure}[t]{0.32\linewidth}
      \centering
      \includegraphics[width=\linewidth]{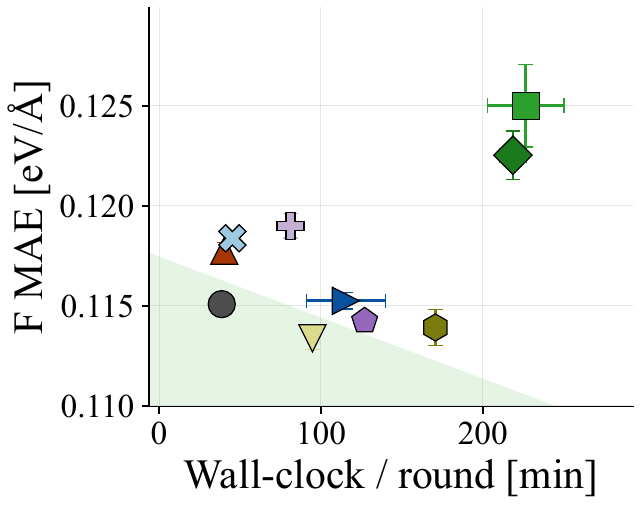}
      \caption{Force MAE vs cost.}
      \label{fig:pareto_oc20}
  \end{subfigure}
  \hfill
  \begin{subfigure}[t]{0.32\linewidth}
      \centering
      \includegraphics[width=\linewidth]{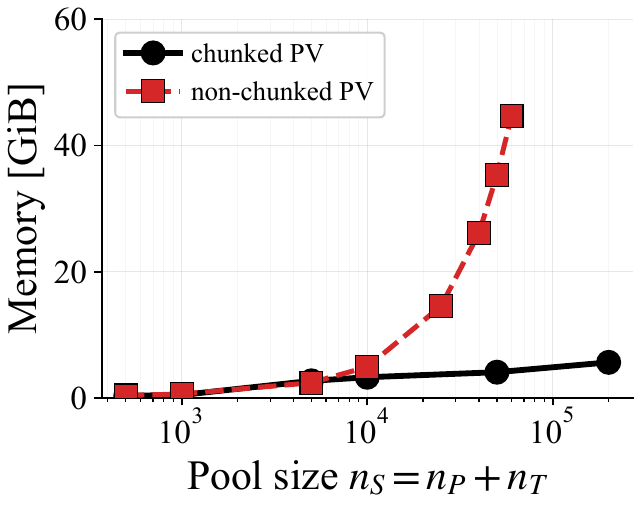}
      \caption{Memory vs.\ pool size.}
      \label{fig:pareto_memory}
  \end{subfigure}
  \caption{Cost, accuracy, and memory scaling of feature-space acquisition on OC20. %Shared legend on top applies to (\subref{fig:pareto_oc20_MAEenergy}) and (\subref{fig:pareto_oc20}).
  (\subref{fig:pareto_oc20_MAEenergy},\subref{fig:pareto_oc20})~Mean per-round wall-clock vs.\ test energy MAE per atom and force MAE at final round. All kernel methods use LCMD selection on a PV shortlist and (reado.) denotes NTK variants on the readouts parameter subset (default is embeddings). The shaded green wedge marks the Pareto region defined through Random and NTK-EF~(reado.). Committee acquisition implementations are not fully optimised, but this does not change the qualitative comparison; we evaluate committees using independently fine-tuned models initialised from the same pretrained checkpoint, requiring \(M+1\) models in total (including a separate evaluation model); excluding this would reduce the reported wall-clock by a factor \(4/3\). In addition, each ensemble member independently rebuilds candidate-pool graphs and executes sequential forward passes, while both energies and forces are computed for all committee variants regardless of the acquisition objective. (\subref{fig:pareto_memory})~Acquisition peak memory of chunked vs.\ non-chunked feature-space PV: the non-chunked baseline grows quadratically, whereas the chunked path grows linearly until the tested \(n_S \approx 2\!\times\!10^5\).}
  \label{fig:pareto_combined}
  \end{figure}
The dominant cost in NTK-based acquisition is the feature-space PV pass over the candidate pool; LCMD selection on the resulting shortlist contributes negligible additional overhead. Two design choices determine the operating point: the parameter subset \(\theta_P\), which fixes the embedding dimension \(d\), and whether the kernel encodes energy, force, or both. Figures~(\subref{fig:pareto_oc20_MAEenergy},\subref{fig:pareto_oc20}) report the cost--accuracy frontier on OC20 across both error axes for energy and force error vs our experiments wall clock time; this additionally includes OC20 runs of the readout NTK variants (see Appendix \ref{app:mace-architecture}). Full per-method runtime and memory breakdowns are reported in Appendix~\ref{app:computational-considerations}. To further characterise the \(\theta_P\) trade-off in isolation, in Appendix \ref{app:timing} we sweep four MACE parameter blocks (readouts, embeddings, interactions, last-layer) on a 5-reaction T1x subset. All NTK subsets achieve similar accuracy, and we find that larger subsets (interaction, last-layer) provide no measurable accuracy gain. 

A further cost constraint is memory during acquisition. A naive non-chunked PV implementation must materialise the kernel matrix \(K \in \mathbb{R}^{n_S \times n_S}\), where $n_{S} = n_{\mathcal{P}} + n_{\mathcal{T}}$, and so grows quadratically in the pool size, whereas chunked feature-space PV maintains only the \(d \times d\) precision matrix and a small chunk buffer (Figure~\ref{fig:pareto_memory}). 
The bottleneck therefore shifts from pool size to feature dimension \(d\), which restricts NTK acquisition to lightweight parameter subsets such as readouts or embeddings.
Moving the bottleneck off pool size matters beyond active learning itself: dataset curation and distillation from million-scale atomistic datasets~\cite{levine2026openmolecules2025omol25,sahoo2025opencatalyst2025oc25} require acquisition signals that scale at most linearly in the candidate set. 

\section{Conclusion}

We address three practical challenges for offline active learning of MLIPs: scaling to large candidate pools, incorporating force-aware acquisition, and maintaining robustness under candidate-pool bias. Our results indicate that these challenges are coupled: scalability widens the operating regime in which the other two matter; force awareness becomes essential as pools grow more chemically heterogeneous; and bias robustness determines whether the resulting acquisition signal remains effective when the candidate pool diverges from the test distribution.

We introduce force-aware NTK based on mixed parameter-coordinate derivatives, together with a joint energy-force NTK that subsumes energy-only and force-only acquisition as special cases. On OC20, the joint kernel is the only method that performs consistently across all in- and out-of-distribution splits, demonstrating that energy-only acquisition is insufficient when force accuracy governs downstream behaviour. On the reactive benchmarks, force-aware variants remain competitive with energy-only baselines but do not uniformly improve over them; characterising the regimes in which force information contributes to acquisition signal energy sensitivity remains an open question. 

We further show that model-based acquisition can be scaled to candidate pools of \(\sim\!200\)k structures with linear complexity in both training-set and pool size via the chunked shortlist-based pipeline. This shifts the dominant computational bottleneck from pool size to feature dimension \(d\), which restricts practical NTK acquisition to lightweight parameter subsets such as embeddings and readouts. Within this regime, the resulting methods are both more accurate and faster than committee-based baselines, since the per-round cost of retraining an ensemble exceeds that of computing feature representations through a single forward and backward pass over the candidate pool. Nevertheless, model-based acquisition is expensive at million-scale candidate sets, where runtime is still dominated by repeated feature extraction over the pool, and fixed-descriptor approaches may offer a cheaper alternative. Further limitations are discussed in Section~\ref{app:limitations}.

We additionally study robustness to candidate-pool bias in a controlled setting on T1x, where reaction-family and reaction-coordinate sampling weights can be varied independently. Across both forms of bias, model-derived representations (both NTK and pretrained activation kernels) produce more stable acquisition behaviour than committee-based disagreement, indicating that an explicit similarity geometry over the pool can dampen the impact of generator-induced bias more effectively than point-wise uncertainty estimates.

There are many directions that follow from this work. The joint energy--force NTK provides a general construction in which physically distinct prediction targets are combined into a single tunable acquisition kernel. A natural extension would be to incorporate additional supervision signals such as stress, Hessians, and partial charges. At the same time, many aspects of the construction remain heuristic, including how best to combine and normalise different representations. More broadly, the offline acquisition setting connects naturally to dataset distillation: the chunked feature-space pipeline introduced here can score and prune candidates from large-scale atomistic datasets, providing a route toward model-aware curation of pretraining and fine-tuning corpora.

\section*{Acknowledgements}
We thank Silvia Acosta Gutiérrez, Massimo Bortone, Heloise Chomet, Valentin Heyraud, Jack Simons, Tamás Lajos Tompa, Lucien Walewski, and Leon Wehrhan for insightful discussions. We also thank Scott Cameron for highlighting the potential usefulness of Neural Tangent Kernels for uncertainty quantification.
\bibliographystyle{plainnat}
\bibliography{main}

\clearpage
\appendix

\section{Limitations}\label{app:limitations}
In this section, we outline further limitations of our work.

Despite the scalability improvements introduced by the chunked feature-space acquisition pipeline, the cost of NTK-based acquisition remains sensitive to the feature dimension \(d\), which restricts practical use to relatively lightweight parameter subsets. Furthermore, the overall runtime is still dominated by repeated feature extraction passes over the candidate pool, which may still become challenging at million-scale candidate sets.

Our force-aware NTK construction also depends on specific aggregation and normalisation choices for atomic force Jacobians and the energy--force mixing parameters. The relative weighting between energy and force contributions remains heuristic and dataset-dependent. The kernel visualisations in Appendix~\ref{app:force_ntk_plot} further suggest that the force contribution can dominate the joint representation, indicating that more sophisticated normalisation or balancing strategies may improve uncertainty estimation.

We focus on offline pool-based active learning and do not evaluate online active-learning workflows in which candidate generation and acquisition are coupled. In addition, all experiments are performed using MACE architectures, and it remains unclear how well the proposed force-aware NTK constructions transfer to other equivariant networks~\cite{schutt2018schnet,batzner2022nequip,wood2026umafamilyuniversalmodels,neumann2025orbv3}.

While we compare against standard committee-based uncertainty estimation, we do not evaluate several recently proposed committee variants designed specifically for pretrained atomistic foundation models~\cite{beck2025multihead}. Additionally, the computational cost of the committee baselines could likely be reduced through more efficient training strategies such as sequential fine-tuning or parameter sharing across ensemble members.

\section{MACE Architecture and Representation Extraction}
\label{app:mace-architecture}

We write a structure as $\mathbf{x}=(\mathbf{z},\mathbf{r})$, with atomic
numbers $\mathbf{z}=(z_i)_{i=1}^{N(\mathbf{x})}$ and Cartesian coordinates
$\mathbf{r}=(\mathbf{r}_i)_{i=1}^{N(\mathbf{x})}$. MACE \cite{batatia2023macehigherorderequivariant} constructs a
radius-cutoff graph $G(\mathbf{x})$: nodes are atoms, and directed edges connect
neighbours within cutoff $r_{\max}$. Node attributes are one-hot species vectors
$\mathbf{a}_i$. For an edge $j\to i$, define the relative vector
$\mathbf{r}_{ji}=\mathbf{r}_j-\mathbf{r}_i$, distance
$d_{ji}=\|\mathbf{r}_{ji}\|$, and direction
$\hat{\mathbf{r}}_{ji}=\mathbf{r}_{ji}/d_{ji}$.

We now outline the relevant parameter subsets of MACE that are used for the NTK and activation representations in the main text. 
\subsection{Relevant MACE parameters}
The embedding stage has two parts: an atomic (node) embedding block and a
radial (edge) embedding block.

\paragraph{Atomic embedding weights.}
The atomic embedding block maps the one-hot species attribute $\mathbf{a}_i$ of
each atom to an initial scalar node feature. Concretely, it is a learnable
lookup table
\begin{equation}
  W^{\mathrm{emb}} \in \mathbb{R}^{Z_{\max}\times c},
  \qquad
  h_{i,L=0}^{(0)} \;=\; W^{\mathrm{emb}}_{z_i,\,:},
\end{equation}
where $Z_{\max}$ is the number of supported atomic species and $c$ is the
number of scalar channels. The initial node feature of atom $i$ is therefore
the row of $W^{\mathrm{emb}}$ indexed by its atomic number $z_i$: it depends
only on the atomic species, not on the local environment, the geometry, or
the rest of the structure. Two atoms of the same species in different
configurations enter the network with identical initial features, and
differences in their downstream representations arise entirely from the
geometric information injected by the interaction blocks. Equivalently, the
embedding weights act as a per-species learnable bias that carries the
chemical identity of the atom into the rest of the model.

\paragraph{Radial embedding.}
The radial embedding block maps each distance $d_{ji}$ to edge features
through a Bessel basis modulated by a smooth cutoff envelope. In the MACE
checkpoints used in this work this basis is parameter-free, so the trainable
parameters of the embedding stage are $W^{\mathrm{emb}}$.
Angular information is injected separately, and not at the embedding stage,
through spherical harmonics of $\hat{\mathbf{r}}_{ji}$.

\paragraph{Interaction block parameters.}
The interaction blocks in MACE construct equivariant messages through
Clebsch--Gordan tensor products between neighbour node features and spherical
harmonics on the edge. Following ~\cite{batatia2023macehigherorderequivariant},
messages along an edge $j \to i$ take the form
\begin{equation}
m^{(\ell)}_{i,k l_3 m_3}
=
\sum_{l_1,m_1,l_2,m_2}
C^{l_3 m_3}_{l_1 m_1,l_2 m_2}
\sum_{j \in \mathcal{N}(i)}
R^{(\ell)}_{k l_1 l_2 l_3}(r_{ji})
Y^{m_2}_{l_2}(\hat{\mathbf r}_{ji})
\sum_{k'}
W^{(\ell)}_{k k' l_2 }
h^{(\ell)}_{j,k' l_2 m_2}.
\end{equation}
where $C^{l_3 m_3}_{l_1 m_1,l_2 m_2}$ are fixed Clebsch--Gordan coefficients,
$R^{(t)}_{k l_1 l_2 l_3}(r_{ji})$ are learned radial interaction weights constructed from the radial embedding, and $W^{(\ell)}_{k k' l_2 }$ is a weight matrix. 

In practice, the interaction $R^{(t)}_{k l_1 l_2 l_3}(r_{ji})$ are parameterized by MLPs acting on the
radial basis embedding of the interatomic distance $r_{ji}$. The interaction
parameter subset used in our NTK ablations consists of these learned radial
tensor-product weights together with the surrounding equivariant linear maps.

\paragraph{Interaction stack.}
Interaction blocks combine neighbour node features with radial and angular
edge features to produce equivariant messages, which are then passed through symmetric tensor products aimed at increasing body order. After the interaction layer
$\ell$, the hidden state of atom $i$ can be written as
\begin{equation}
  h_i^{(\ell)}
  =
  \bigoplus_{L=0}^{L_{\max}} h_{i,L}^{(\ell)},
\end{equation}
where $L=0$ channels are invariant scalars and $L>0$ channels transform as
vectors or higher-order tensors under rotations.

\paragraph{Readout parameters.}

After each interaction layer, MACE maps the invariant ($L=0$) node features
to per-site energy contributions through equivariant readout functions, the contribution from interaction layer $t$ is
\begin{equation}
E_i^{(t)}
=
\mathcal R_t\!\left(h_i^{(t)}\right)
=
\begin{cases}
\sum_{\tilde{k}}
W^{(t)}_{\mathrm{readout},\tilde{k}}
\, h^{(t)}_{i,\tilde{k}00},
& t < T, \\[6pt]
\mathrm{MLP}^{(t)}_{\mathrm{readout}}
\!\left(
\left\{
h^{(t)}_{i,\tilde{k}00}
\right\}_{\tilde{k}}
\right),
& t = T.
\end{cases}
\end{equation}
Here $h^{(t)}_{i,\tilde{k}00}$ denotes the invariant scalar channels of the
equivariant node representation. The total energy then has the form
\begin{equation}
  E_\theta(\mathbf{x})
  =
  \sum_{i=1}^{N(\mathbf{x})}
  \sum_{\ell=0}^{T}
  E_i^{(\ell)}(\mathbf{x}),
\end{equation}

and forces are analytic coordinate gradients,
\begin{equation}
  \mathbf{F}_i(\mathbf{x}) = -\nabla_{\mathbf{r}_i} E_\theta(\mathbf{x}).
\end{equation}

The readout parameters therefore consist of
the linear projection weights
$W^{(t)}_{\mathrm{readout},\tilde{k}}$
used in intermediate layers together with the final MLP readout applied at the
last interaction layer. Since the readout depends only on invariant channels,
the resulting site energies are rotationally invariant.

\subsection{Representation extraction.}
We extract two classes of representations from the pretrained MACE model:
activation features and NTK features. 

For both force and energy NTK representations, we differentiate the predicted energy (or force) with respect to
a chosen parameter subset $\theta_P$.
In this work, we consider three choices of $\theta_P$:

The \textit{embedding NTK} uses
$\theta_P = W^{\mathrm{emb}}$, so its features measure the sensitivity of the
energy to the learned species embeddings. Although these parameters are
species-indexed, the gradients are backpropagated through the full interaction
stack and therefore depend on both composition and geometry. 

The \textit{interaction NTK} uses the parameters of the equivariant interaction blocks, including the
learned radial tensor-product weights
$R^{(\ell)}_{k l_1 l_2 l_3}$ and the associated equivariant linear maps. This
subset captures sensitivity to the message-passing operations that couple local
geometry, angular information, and neighbour features. 

The \textit{readout NTK} uses the
parameters of the scalar readout functions, including the linear readout weights
$W^{(t)}_{\mathrm{readout},\tilde{k}}$ and the final readout MLP. This subset
acts only on invariant scalar channels and is therefore lower-dimensional and
cheaper to compute than the interaction NTK.

Unless otherwise stated, the NTK representation in our work uses the embedding-parameter subset, while the cost-accuracy
analysis additionally compares embedding, interaction, and readout subsets in Section \ref{sec:scaling}.

The activation representation instead pools the
invariant scalar node features from each message-passing layer and concatenates
them,
\begin{equation}
\phi_{\mathrm{act}}(\mathbf{x})
=
\mathrm{concat}
(
\{
\sum_{i=1}^{N(\mathbf{x})}
h^{(\ell)}_{i,L=0}
\}_{\ell=1}^{T}
),
\end{equation} This representation uses
the hidden node features directly and does not involve differentiating with
respect to model parameters.

\section{Training Details}
\label{app:training-details}

Active-learning rounds fine-tune pretrained MACE models. Model weights are optimised on the current labelled set without freezing any model parameters. At each round, the MLIP model is initialised from the pretrained weights and is fine-tuned on the augmented dataset. This is commonly done in active learning as fine-tuning on only the newly acquired data causes catastrophic forgetting of data acquired at previous rounds. The surrogate predicts total energies and atomic forces, and the
training objective is a Huber loss on energy and force errors with equal energies and force weights
$\lambda_E=\lambda_F=10.0$. We use a variety of settings in order to obtain converged finetunings for the variety of dataset sizes and scenarios we train on. 

\paragraph{Reactivity datasets:}
For the T1x, RGD, PMechDB active learnings in Section \ref{sec:force_ntk_main}, the batch size and learning rate are chosen as
\begin{equation}
(B,\eta)
=
\begin{cases}
(4,\,10^{-2}),
& |T^{(t)}| \leq 250, \\[4pt]
(16,\,10^{-2}),
& |T^{(t)}| > 250,
\end{cases}
\end{equation}
where $|T^{(t)}|$ denotes the size of the labelled training set at active-learning
round $t$. The number of epochs is chosen as
\begin{equation}
E
=
\max\left(
50,
\left\lceil
\frac{
4000 \, B
}{
|T^{(t)}|
}
\right\rceil
\right).
\end{equation}

\paragraph{Small T1x datasets:}
For the small T1x dataset trainings (the biasing in Section \ref{sec:cpshift}, and cost analysis \ref{sec:scaling}), we start off with extremely small training sets. Each round therefore uses a dynamic schedule~\cite{pretrained}. If
$|\mathcal{T}^{(t)}|$ is the current labelled-set size, the batch size $B$ and
learning rate $\eta$ are
\begin{equation}
    (B,\eta)=
\begin{cases}
(1,10^{-3}), & |\mathcal{T}^{(t)}|\le 20,\\
(2,5\times 10^{-3}), & 20<|\mathcal{T}^{(t)}|\le 100,\\
(4,5\times 10^{-3}), & |\mathcal{T}^{(t)}|>100,
\end{cases}
\end{equation}

and the number of epochs is
\begin{equation}
    E=\max\!\left(10,\left\lceil\frac{1000B}{|\mathcal{T}^{(t)}|}\right\rceil\right).
\end{equation}

\paragraph{OC20:}
For OC20, we use a fixed batch size and
learning rate,
\begin{equation}
(B,\eta)
=
(4,\,10^{-3}),
\end{equation}
together with a dynamically adjusted number of epochs
\begin{equation}
E
=
\max\left(
30,
\left\lceil
\frac{
1000 \times B
}{
|T^{(t)}|
}
\right\rceil
\right),
\end{equation}
where $|T^{(t)}|$ denotes the size of the labelled training set at active-learning
round $t$.

\clearpage

\subsection{Pretrained models}
The hyperparameters from the SPICE2 and MPTraj models can be found in Tables \ref{tab:spice_hyperparameters} and \ref{tab:mptraj_hyperparameters}
\begin{table}[h]
\centering
\caption{Hyperparameters for the SPICE 2 MACE model.}
\label{tab:spice_hyperparameters}
\begin{tabular}{ll}
\hline
\textbf{Hyperparameter} & \textbf{Value} \\
\hline
Number of interaction layers ($N_{\mathrm{layers}}$) & 2 \\
Number of channels ($N_{\mathrm{channels}}$) & 128 \\
Maximum angular momentum ($\ell_{\max}$) & 3 \\
Node symmetry & 3 \\
Correlation order & 2 \\
Readout irreducible representations & $\{16 \times 0e,\;0e\}$ \\
Include pseudotensors & False \\
Number of Bessel basis functions & 8 \\
Activation function & SiLU \\
Radial envelope & Polynomial envelope \\
Cutoff distance (\AA)  & $5.0~\mathrm{\AA}$ \\
Number of species & 15 \\
\hline
\end{tabular}
\end{table}

\begin{table}[h]
\centering
\caption{Hyperparameters of the MPTraj MACE model.}
\label{tab:mptraj_hyperparameters}
\begin{tabular}{ll}
\hline
\textbf{Hyperparameter} & \textbf{Value} \\
\hline
Number of interaction layers ($N_{\mathrm{layers}}$) & 2 \\
Number of channels ($N_{\mathrm{channels}}$) & 128 \\
Maximum angular momentum ($\ell_{\max}$) & 2 \\
Node symmetry & 2 \\
Correlation order & 3 \\
Readout irreducible representations & $\{16 \times 0e,\;0e\}$ \\
Number of Bessel basis functions & 10 \\
Include pseudotensors & False \\
Activation function & SiLU \\
Radial envelope & Polynomial envelope \\
Cutoff distance (\AA) & $6.0~\mathrm{\AA}$ \\
Number of species & 88 \\
\hline
\end{tabular}
\end{table}
\section{Scalable feature-space acquisition}
\label{app:scalable_pv}

This section details the shortlist acquisition pipeline used for large candidate pools. The implementation operates entirely in feature space and scales linearly in the pool and train size due to the shortlisting procedure.
\subsection{Feature-space posterior variance}
In this section, for completeness, we give a derivation that posterior variance acquisition is equivalent to scoring via the feature matrix $M_{\mathcal{T}}$ \cite{Rasmussen2005}.

For kernels induced by a feature map \(k(\mathbf{x}, \mathbf{x}') = \phi(\mathbf{x})^\top \phi(\mathbf{x}')\) with \(\phi(\mathbf{x}) \in \mathbb{R}^d\), posterior variance can be computed without forming the kernel matrix. Let the training feature matrix be \(\Phi_{\mathcal{T}} \in \mathbb{R}^{n_{\mathcal{T}} \times d}\) and the candidate pool size be \(n_{\mathcal{P}}\). The Gaussian posterior variance for a candidate \(\mathbf{x}\) is
\begin{equation}
\sigma_{\mathcal{T}}^2(\mathbf{x})
=
k(\mathbf{x},\mathbf{x})
-
k_{\mathcal{T}}(\mathbf{x})^\top
\left(k_{\mathcal{T}\mathcal{T}} + \lambda I\right)^{-1}
k_{\mathcal{T}}(\mathbf{x}),
\end{equation}
where \(k_{\mathcal{T}\mathcal{T}} = \Phi_{\mathcal{T}} \Phi_{\mathcal{T}}^\top\) and \(k_{\mathcal{T}}(\mathbf{x}) = \Phi_{\mathcal{T}} \phi(\mathbf{x})\). Substituting gives
\begin{equation}
\sigma_{\mathcal{T}}^2(\mathbf{x})
=
\phi(\mathbf{x})^\top
\left[
I - \Phi_{\mathcal{T}}^\top
\left(\Phi_{\mathcal{T}} \Phi_{\mathcal{T}}^\top + \lambda I\right)^{-1}
\Phi_{\mathcal{T}}
\right]
\phi(\mathbf{x}).
\end{equation}
Applying the Woodbury identity \cite{Golub2012},
\begin{equation}
    I - \Phi_{\mathcal{T}}^\top(\Phi_{\mathcal{T}}\Phi_{\mathcal{T}}^\top + \lambda I)^{-1}\Phi_{\mathcal{T}}
= \lambda(\Phi_{\mathcal{T}}^\top \Phi_{\mathcal{T}} + \lambda I)^{-1},
\end{equation}

yields
\begin{equation}
\sigma_{\mathcal{T}}^2(\mathbf{x})
=
\lambda\, \phi(\mathbf{x})^\top
(\Phi_{\mathcal{T}}^\top \Phi_{\mathcal{T}} + \lambda I)^{-1}
\phi(\mathbf{x}).
\end{equation}
Defining the feature-space precision matrix \(M_{\mathcal{T}} = (\Phi_{\mathcal{T}}^\top \Phi_{\mathcal{T}} + \lambda I)^{-1} \in \mathbb{R}^{d \times d}\), we obtain \(\sigma_{\mathcal{T}}^2(\mathbf{x}) = \lambda\, \phi(\mathbf{x})^\top M_{\mathcal{T}} \phi(\mathbf{x})\). Since \(\lambda\) is constant, ranking by posterior variance is equivalent to scoring by \(s(\mathbf{x}) = \phi(\mathbf{x})^\top M_{\mathcal{T}} \phi(\mathbf{x})\). 

To compute $M_{\mathcal{T}}$ efficiently, we accumulate the Gram matrix \(\Phi_{\mathcal{T}}^\top \Phi_{\mathcal{T}}\) in chunks. Specifically, for chunks \(\Phi_C \in \mathbb{R}^{c \times d}\) of the training features, we maintain a running sum $\sum_C \Phi_C^\top \Phi_C$ after which the feature precision matrix can be formed. This requires materialising matrices of size at most \(O(\max(cd, d^2))\), making the computation linear in \(n_{\mathcal{T}}\) and memory-efficient in the number of training points. Candidate scores \(s(\mathbf{x})\) for \(\mathbf{x} \in \mathcal{P}\) are then also computed in chunks using the same strategy. During the scoring pass, we maintain a running top-\(K\) representation shortlist by caching the highest-scoring candidates, enabling efficient selection without evaluating or storing the full pool.

\subsection{LCMD}
Posterior variance alone tends to favour isolated high-uncertainty points. We therefore use the posterior variance score only to shortlist a set of \(K\) candidates with the highest uncertainty. We then perform batch selection using Largest-Cluster Maximum-Distance (LCMD)~\cite{holzmuller2023batch_al} on the shortlisted set.

LCMD modifies greedy farthest-point sampling to favour dense but under-covered regions. Let \(S\) denote the set of points already selected into the current batch. At each step, every remaining candidate is assigned to its nearest centre in \(S \cup \mathcal{T}^{(t)}\). For each resulting cluster \(\mathcal{C}_c\) with centre \(\mathbf{z}_c\), we compute the total squared distance mass
\begin{equation}
m_c
=
\sum_{\mathbf{x} \in \mathcal{C}_c}
d_k^2(\mathbf{x},\mathbf{z}_c),
\end{equation}
where the kernel-induced squared distance is
\begin{equation}
d_k^2(\mathbf{x},\mathbf{z})
= \|\phi(\mathbf{x}) - \phi(\mathbf{z})\|_2^2.
\end{equation}
LCMD then selects the cluster with the largest mass and adds the point within that cluster farthest from its centre:
\begin{equation}
\mathbf{x}_{\mathrm{next}}
=
\arg\max_{\mathbf{x}\in\mathcal{C}_{c^\star}}
d_k^2(\mathbf{x},\mathbf{z}_{c^\star}),
\qquad
c^\star = \arg\max_c m_c.
\end{equation}

By using shortlisting, the cost of including diversity via LCMD is changed from  $O\!\left(n_{\mathcal{P}} n_{\mathcal{T}}\right)$ to $O\!\left(K n_{\mathcal{T}}\right)$. Pseudo code for the full shortlisting is given in Algorithm \ref{alg:chunked_pv} and detailed performance summaries can be found in Appendix \ref{app:computational-considerations}.

\begin{algorithm}[h]
\caption{Chunked feature-space PV with shortlist selection}
\label{alg:chunked_pv}
\begin{algorithmic}[1]
\REQUIRE Feature map \(\phi\), training set \(\mathcal{T}\), pool \(\mathcal{P}\), batch size \(B\), shortlist size \(K\), chunk size \(c\), regularisation \(\lambda\)
\STATE Initialize \(G_{\mathcal{T}} \leftarrow 0_{d \times d}\)
\FOR{chunks \(\mathcal{C}_{\mathcal{T}} \subset \mathcal{T}\) of size \(\le c\)}
    \STATE Compute features \(\Phi_{\mathcal{C}_{\mathcal{T}}} = \phi(\mathcal{C}_{\mathcal{T}}) \in \mathbb{R}^{|\mathcal{C}_{\mathcal{T}}| \times d}\)
    \STATE Accumulate \(G_{\mathcal{T}} \leftarrow G_{\mathcal{T}} + \Phi_{\mathcal{C}_{\mathcal{T}}}^{\top}\Phi_{\mathcal{C}_{\mathcal{T}}}\)
\ENDFOR
\STATE Compute \(M_{\mathcal{T}} = (G_{\mathcal{T}} + \lambda I)^{-1}\)
\STATE Initialize empty top-\(K\) shortlist \(\mathcal{S}\)
\FOR{chunks \(\mathcal{C} \subset \mathcal{P}\) of size \(\le c\)}
    \STATE Compute features \(\Phi_{\mathcal{C}} = \phi(\mathcal{C}) \in \mathbb{R}^{|\mathcal{C}| \times d}\)
    \STATE Compute scores \(s_{\mathcal{C}} = \operatorname{diag}(\Phi_{\mathcal{C}} M_{\mathcal{T}} \Phi_{\mathcal{C}}^\top)\)
    \STATE Update \(\mathcal{S}\) with the top-\(K\) candidates seen so far
\ENDFOR
\STATE Run LCMD or greedy PV on \(\mathcal{S}\) to select batch \(\mathcal{A}\) of size \(B\)
\RETURN \(\mathcal{A}\)
\end{algorithmic}
\end{algorithm}

\clearpage
\section{Additional reactivity information}\label{app:reactivity}
In this section, we give further plots and benchmarks for Section \ref{sec:force_ntk_main} of the main text. 

\subsection{MAE metrics}
Table~\ref{tab:reactive_summary_mae} reports the MAE area under the learning curve together with the final energy and force MAE values. Compared to the RMSE results in Table~\ref{tab:reactive_summary} of the main text, the improvements under active learning are generally smaller for MAE than for RMSE. This is consistent with active learning preferentially reducing high-error regions of the distribution, since RMSE is more sensitive to rare but large errors. We additionally observe that committee-based methods exhibit a substantially larger gap between RMSE and MAE improvements than the representation-based methods. This suggests that a significant fraction of their RMSE gains may arise from preferential selection of outlier or high-error structures, potentially due to the lack of explicit conditioning on the training set or candidate-pool geometry.
\begin{table}[h]
\centering
\scriptsize
\setlength{\tabcolsep}{2pt}

\caption{
Energy and force acquisition performance across reactive benchmarks using MAE metrics.
Reported metrics are energy MAE area under the acquisition curve (AUC) and final-round MAE in meV, and force MAE AUC and final-round MAE in meV/\AA.
Lower is better; best values in each row are bolded.
Percent differences are relative to Random.
}

\label{tab:reactive_summary_mae}

\begin{tabular}{llccccccc}
\toprule
\textbf{Dataset} & \textbf{Metric} &
\textbf{Random} &
\textbf{Activation} &
\textbf{Committee-E} &
\textbf{Committee-F} &
\textbf{NTK-E} &
\textbf{NTK-F} &
\textbf{NTK-EF} \\
\midrule

\multirow{4}{*}{\textbf{RGD}}

& E AUC
& 193.5
& \textbf{189.0} \imp{-2.3\%}
& 202.5 \worse{+4.7\%}
& 210.0 \worse{+8.5\%}
& 192.5 \imp{-0.5\%}
& 189.0 \imp{-2.3\%}
& 191.5 \imp{-1.0\%} \\

& F AUC
& \textbf{2170.0}
& 2185.0 \worse{+0.7\%}
& 2237.0 \worse{+3.1\%}
& 2252.5 \worse{+3.8\%}
& 2176.0 \worse{+0.3\%}
& 2174.0 \worse{+0.2\%}
& 2171.5 \worse{+0.1\%} \\

& Final E 
& 6.5
& \textbf{6.0} \imp{-7.7\%}
& 6.5 (0.0\%)
& \textbf{6.0} \imp{-7.7\%}
& \textbf{6.0} \imp{-7.7\%}
& \textbf{6.0} \imp{-7.7\%}
& 6.5 (0.0\%) \\

& Final F 
& 85.5
& 85.0 \imp{-0.6\%}
& 85.5 (0.0\%)
& \textbf{84.5} \imp{-1.2\%}
& 85.0 \imp{-0.6\%}
& \textbf{84.5} \imp{-1.2\%}
& 85.0 \imp{-0.6\%} \\

\midrule

\multirow{4}{*}{\textbf{T1X}}

& E AUC
& 90.5
& \textbf{70.5} \imp{-22.1\%}
& 155.0 \worse{+71.3\%}
& 143.0 \worse{+58.0\%}
& 73.0 \imp{-19.3\%}
& 75.5 \imp{-16.6\%}
& 75.0 \imp{-17.1\%} \\

& F AUC
& 843.0
& 700.0 \imp{-16.9\%}
& 1082.0 \worse{+28.4\%}
& 990.5 \worse{+17.5\%}
& \textbf{697.0} \imp{-17.3\%}
& 711.5 \imp{-15.6\%}
& 703.5 \imp{-16.5\%} \\

& Final E 
& 2.0
& \textbf{1.5} \imp{-25.0\%}
& 2.0 (0.0\%)
& 2.0 (0.0\%)
& 2.0 (0.0\%)
& 2.0 (0.0\%)
& 2.0 (0.0\%) \\

& Final F 
& 20.5
& \textbf{19.0} \imp{-7.3\%}
& 20.5 (0.0\%)
& 20.0 \imp{-2.4\%}
& 19.5 \imp{-4.9\%}
& 19.5 \imp{-4.9\%}
& 20.0 \imp{-2.4\%} \\

\midrule

\multirow{4}{*}{\textbf{PMechDB}}

& E MAE 
& 316.0
& \textbf{305.0} \imp{-3.5\%}
& 352.0 \worse{+11.4\%}
& 334.5 \worse{+5.9\%}
& 307.5 \imp{-2.7\%}
& 311.0 \imp{-1.6\%}
& 313.0 \imp{-0.9\%} \\

& F MAE 
& \textbf{1061.5}
& 1074.5 \worse{+1.2\%}
& 1104.0 \worse{+4.0\%}
& 1116.5 \worse{+5.2\%}
& 1072.5 \worse{+1.0\%}
& 1064.0 \worse{+0.2\%}
& 1068.0 \worse{+0.6\%} \\

& Final E 
& 11.5
& \textbf{11.0} \imp{-4.3\%}
& 13.0 \worse{+13.0\%}
& 12.0 \worse{+4.3\%}
& \textbf{11.0} \imp{-4.3\%}
& \textbf{11.0} \imp{-4.3\%}
& 11.5 (0.0\%) \\

& Final F 
& 41.5
& 41.5 (0.0\%)
& 42.0 \worse{+1.2\%}
& 42.0 \worse{+1.2\%}
& \textbf{41.0} \imp{-1.2\%}
& \textbf{41.0} \imp{-1.2\%}
& \textbf{41.0} \imp{-1.2\%} \\

\bottomrule
\end{tabular}
\end{table}

\subsection{Descriptor baselines}

\begin{table}[h]
\centering
\scriptsize
\setlength{\tabcolsep}{3pt}

\caption{
RMSE-based comparison of Tanimoto, SOAP, and NTK-EF acquisition strategies across reactive benchmarks.
Reported values include energy RMSE area under the acquisition curve (AUC) and final-round errors in meV, together with force RMSE AUC and final-round errors in meV/\AA.
Percent differences are relative to the Random baseline.
}

\label{tab:tanimoto_soap_ntkef_rmse}

\begin{tabular}{llccc}
\toprule
\textbf{Dataset} & \textbf{Metric}
& \textbf{Tanimoto}
& \textbf{SOAP}
& \textbf{NTK-EF} \\
\midrule

\multirow{4}{*}{RGD}
& E AUC
& 339.0 \imp{-2.6\%}
& \textbf{338.5} \imp{-2.7\%}
& 344.5 \imp{-1.0\%} \\

& F AUC
& 4735.5 \imp{-0.6\%}
& 4744.0 \imp{-0.4\%}
& \textbf{4684.0} \imp{-1.7\%} \\

& Final E
& \textbf{12.0} \imp{-7.7\%}
& \textbf{12.0} \imp{-7.7\%}
& \textbf{12.0} \imp{-7.7\%} \\

& Final F
& 188.0 \imp{-0.3\%}
&\textbf{186.0} \imp{-1.3\%}
& \textbf{186.0} \imp{-1.3\%} \\

\midrule

\multirow{4}{*}{T1X}
& E AUC
& 189.5 \imp{-23.1\%}
& \textbf{171.5} \imp{-30.4\%}
& 175.0 \imp{-29.0\%} \\

& F AUC
& 2008.5 \imp{-12.7\%}
& 1914.0 \imp{-16.8\%}
& \textbf{1796.5} \imp{-21.9\%} \\

& Final E
& 6.0 \imp{-7.7\%}
& 5.5 \imp{-15.4\%}
& \textbf{5.0} \imp{-23.1\%} \\

& Final F
& 61.5 \imp{-0.8\%}
& \textbf{58.0} \imp{-6.5\%}
& 60.5 \imp{-2.4\%} \\

\midrule

\multirow{4}{*}{PMechDB}
& E AUC
& 450.0 \imp{-0.7\%}
& 442.0 \imp{-2.4\%}
& \textbf{431.0} \imp{-4.9\%} \\

& F AUC
& \textbf{2739.5} \imp{-4.1\%}
& 2993.5 \worse{+4.8\%}
& 2777.0 \imp{-2.8\%} \\

& Final E
& 17.0 (0.0\%)
& \textbf{16.5} \imp{-2.9\%}
& \textbf{16.5} \imp{-2.9\%} \\

& Final F
& 111.0 \imp{-2.2\%}
& 119.5 \worse{+5.3\%}
& \textbf{109.5} \imp{-3.5\%} \\

\bottomrule
\end{tabular}
\end{table}
\begin{table}[h]
\centering
\scriptsize
\setlength{\tabcolsep}{3pt}

\caption{
MAE-based comparison of Tanimoto, SOAP, and NTK-EF acquisition strategies across reactive benchmarks.
Reported values include energy MAE area under the acquisition curve (AUC) and final-round errors in meV, together with force MAE AUC and final-round errors in meV/\AA.
Percent differences are relative to the Random baseline.
}

\label{tab:tanimoto_soap_ntkef_mae}

\begin{tabular}{llccc}
\toprule
\textbf{Dataset} & \textbf{Metric}
& \textbf{Tanimoto}
& \textbf{SOAP}
& \textbf{NTK-EF} \\
\midrule

\multirow{4}{*}{RGD}
& E AUC
& 188.5 \imp{-2.6\%}
& 192.0 \imp{-0.8\%}
& 191.5 \imp{-1.0\%} \\

& F AUC
& 2170.0 (0.0\%)
& 2191.0 \worse{+1.0\%}
& 2171.5 \worse{+0.1\%} \\

& Final E
& 6.5 (0.0\%)
& 6.5 (0.0\%)
& 6.5 (0.0\%) \\

& Final F
& 85.5 (0.0\%)
& 85.0 \imp{-0.6\%}
& 85.0 \imp{-0.6\%} \\

\midrule

\multirow{4}{*}{T1X}
& E AUC
& 82.0 \imp{-9.4\%}
& 74.0 \imp{-18.2\%}
& 75.0 \imp{-17.1\%} \\

& F AUC
& 774.5 \imp{-8.1\%}
& 735.0 \imp{-12.8\%}
& 703.5 \imp{-16.5\%} \\

& Final E
& 2.0 (0.0\%)
& 1.5 \imp{-25.0\%}
& 2.0 (0.0\%) \\

& Final F
& 21.0 \worse{+2.4\%}
& 20.0 \imp{-2.4\%}
& 20.0 \imp{-2.4\%} \\

\midrule

\multirow{4}{*}{PMechDB}
& E AUC
& 322.5 \worse{+2.1\%}
& 312.5 \imp{-1.1\%}
& 313.0 \imp{-0.9\%} \\

& F AUC
& 1044.5 \imp{-1.6\%}
& 1089.5 \worse{+2.6\%}
& 1068.0 \worse{+0.6\%} \\

& Final E
& 12.0 \worse{+4.3\%}
& 11.0 \imp{-4.3\%}
& 11.5 (0.0\%) \\

& Final F
& 41.0 \imp{-1.2\%}
& 42.5 \worse{+2.4\%}
& 41.0 \imp{-1.2\%} \\

\bottomrule
\end{tabular}
\end{table}

\clearpage
\section{Energy-force weighting sweep}
Here we display results as the relative weighting \(w_E{:}w_F\) is varied. Across the reactive benchmarks, we generally observe small improvements when increasing the relative energy weighting, supporting the conclusion that energy-based acquisition already captures much of the informative acquisition signal for these datasets. Overall, however, the differences between weighting variants remain small and are often attributable to only a handful of selected structures across the active-learning trajectory. Although the underlying kernels are linear combinations of energy and force components, the downstream acquisition procedures are not, so even small changes in kernel geometry can lead to different candidate selections over successive AL rounds.
\begin{table*}[h]
\centering
\scriptsize
\setlength{\tabcolsep}{3pt}

\caption{
Comparison of NTK-based acquisition strategies with different energy-force weighting ratios across reactive benchmarks.
Reported values include energy/force RMSE area under the acquisition curve (AUC) together with final-round errors.
Lower is better; best values in each row are bolded. Units are in meV and meV/\AA. Percent differences are relative to the Random baseline.
}

\label{tab:ntk_weighting_rmse}

\begin{tabular}{llcccccc}
\toprule
\textbf{Dataset} & \textbf{Metric}
& \textbf{NTK-E}
& \textbf{NTK-F}
& \textbf{NTK-EF 1:1}
& \textbf{NTK-EF 4:1}
& \textbf{NTK-EF 32:1}
& \textbf{NTK-EF 128:1} \\
\midrule

\multirow{4}{*}{RGD}

& E AUC
& 339.0 \imp{-2.6\%}
& 338.0 \imp{-2.9\%}
& 344.5 \imp{-1.0\%}
& 341.5 \imp{-1.9\%}
& 330.0 \imp{-5.2\%}
& \textbf{323.0} \imp{-7.2\%} \\

& F AUC
& 4651.5 \imp{-2.4\%}
& 4665.5 \imp{-2.1\%}
& 4684.0 \imp{-1.7\%}
& 4656.0 \imp{-2.3\%}
& \textbf{4614.5} \imp{-3.1\%}
& 4638.0 \imp{-2.6\%} \\

& Final E
& 12.0 \imp{-7.7\%}
& 12.0 \imp{-7.7\%}
& 12.0 \imp{-7.7\%}
& 12.0 \imp{-7.7\%}
& 12.0 \imp{-7.7\%}
& 12.0 \imp{-7.7\%} \\

& Final F
& 186.0 \imp{-1.3\%}
& \textbf{184.0} \imp{-2.4\%}
& 186.0 \imp{-1.3\%}
& 186.0 \imp{-1.3\%}
& \textbf{184.0} \imp{-2.4\%}
& 185.5 \imp{-1.6\%} \\

\midrule

\multirow{4}{*}{T1X}

& E AUC
& \textbf{159.5} \imp{-35.3\%}
& 178.5 \imp{-27.6\%}
& 175.0 \imp{-29.0\%}
& 168.5 \imp{-31.6\%}
& 167.5 \imp{-32.0\%}
& 162.5 \imp{-34.1\%} \\

& F AUC
& \textbf{1754.5} \imp{-23.7\%}
& 1802.5 \imp{-21.6\%}
& 1796.5 \imp{-21.9\%}
& 1775.0 \imp{-22.8\%}
& 1757.5 \imp{-23.6\%}
& 1770.0 \imp{-23.0\%} \\

& Final E
& \textbf{5.0} \imp{-23.1\%}
& 5.5 \imp{-15.4\%}
& \textbf{5.0} \imp{-23.1\%}
& 5.5 \imp{-15.4\%}
& \textbf{5.0} \imp{-23.1\%}
& 5.5 \imp{-15.4\%} \\

& Final F
& \textbf{57.0} \imp{-8.1\%}
& 59.5 \imp{-4.0\%}
& 60.5 \imp{-2.4\%}
& 59.5 \imp{-4.0\%}
& 58.0 \imp{-6.5\%}
& 58.0 \imp{-6.5\%} \\

\midrule

\multirow{4}{*}{PMechDB}

& E AUC
& 426.5 \imp{-5.9\%}
& 433.5 \imp{-4.3\%}
& 431.0 \imp{-4.9\%}
& 432.5 \imp{-4.5\%}
& 427.0 \imp{-5.7\%}
& \textbf{423.5} \imp{-6.5\%} \\

& F AUC
& \textbf{2718.5} \imp{-4.9\%}
& 2780.5 \imp{-2.7\%}
& 2777.0 \imp{-2.8\%}
& 2739.0 \imp{-4.1\%}
& 2777.5 \imp{-2.8\%}
& 2777.5 \imp{-2.8\%} \\

& Final E
& \textbf{16.0} \imp{-5.9\%}
& \textbf{16.0} \imp{-5.9\%}
& 16.5 \imp{-2.9\%}
& 16.5 \imp{-2.9\%}
& \textbf{16.0} \imp{-5.9\%}
& \textbf{16.0} \imp{-5.9\%} \\

& Final F
& \textbf{106.5} \imp{-6.2\%}
& 111.5 \imp{-1.8\%}
& 109.5 \imp{-3.5\%}
& 109.0 \imp{-4.0\%}
& 110.5 \imp{-2.6\%}
& 109.5 \imp{-3.5\%} \\

\bottomrule
\end{tabular}
\end{table*}
\clearpage
\subsection{Learning curves}
This section displays the learning curves for energy and force RMSE/MAE across the reactive benchmarks. Curves are averaged over two seeds. Energy errors are reported in meV, while force errors are reported in meV/\AA.

\begin{figure}[h]
  \centering
  \includegraphics[width=\linewidth]{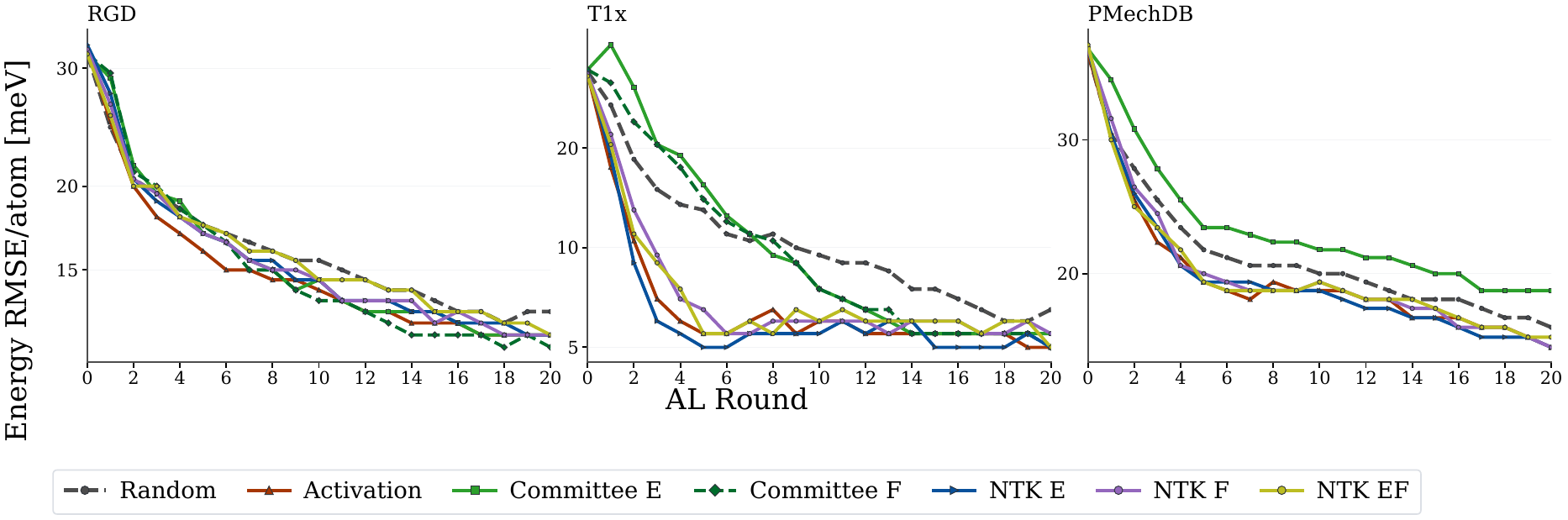}
  \caption{Energy RMSE learning curves (meV) for the methods presented in the main text.}
  \label{fig:rmse_e}
\end{figure}

\begin{figure}[h]
  \centering
  \includegraphics[width=\linewidth]{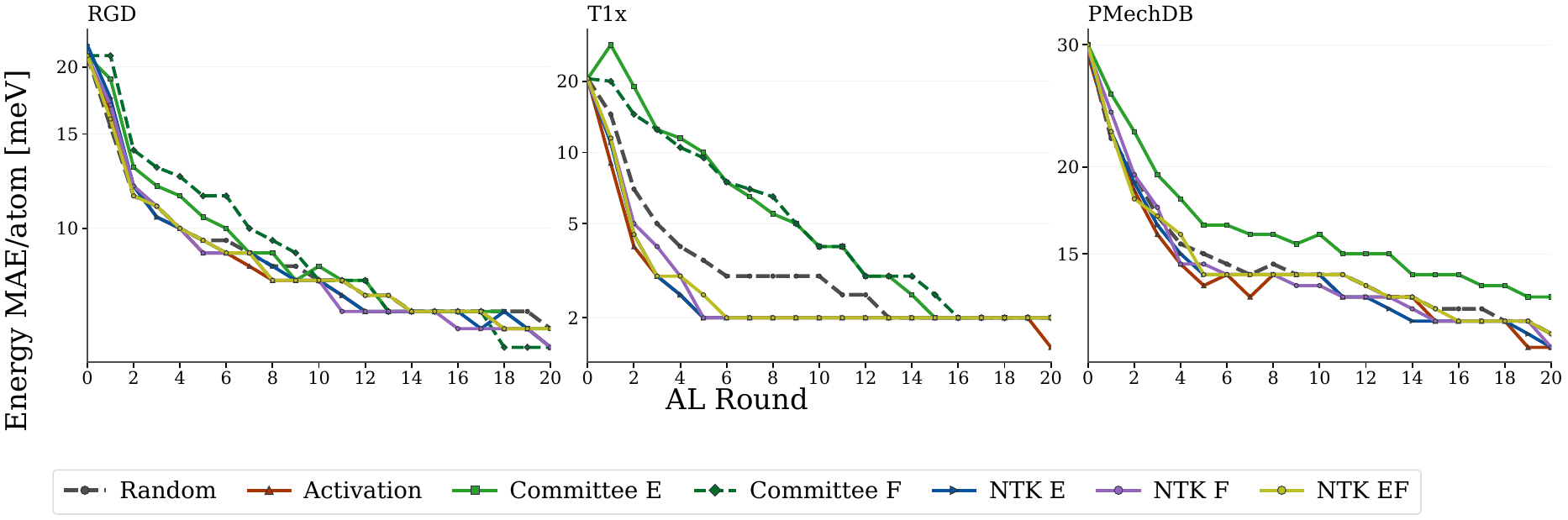}
  \caption{Energy MAE learning curves (meV) for the methods presented in the main text.}
  \label{fig:mae_e}
\end{figure}

\begin{figure}[h]
  \centering
  \includegraphics[width=\linewidth]{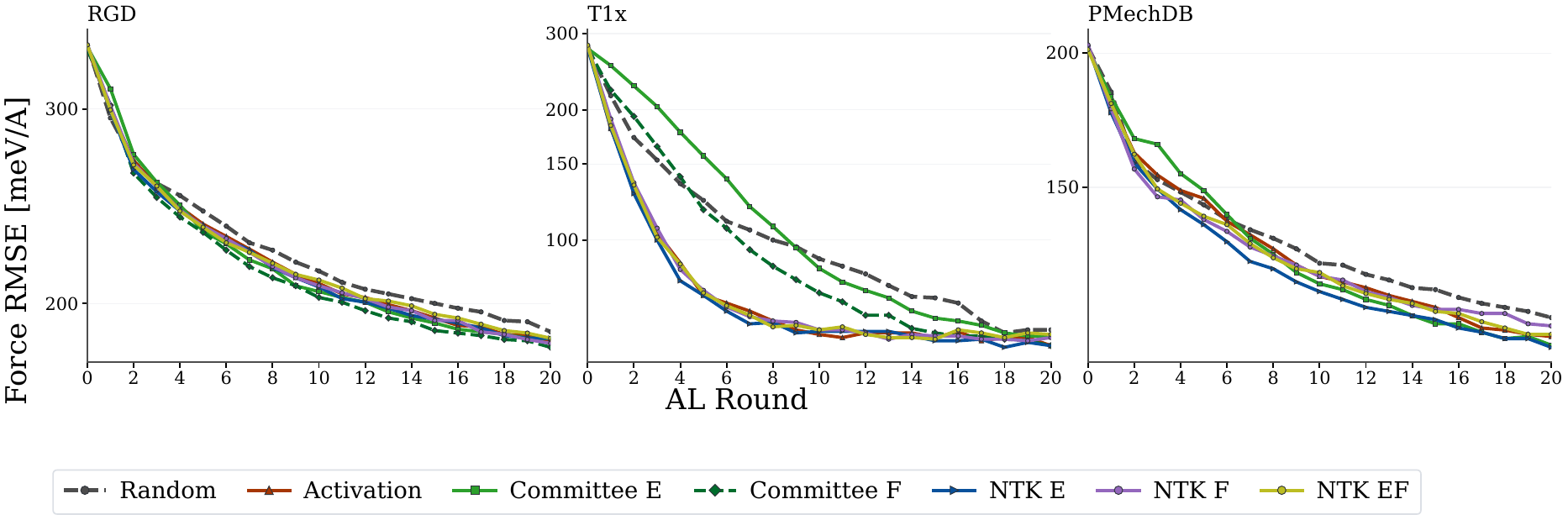}
  \caption{Force RMSE learning curves (meV/\AA) for the methods presented in the main text.}
  \label{fig:rmse_f}
\end{figure}

\begin{figure}[h]
  \centering
  \includegraphics[width=\linewidth]{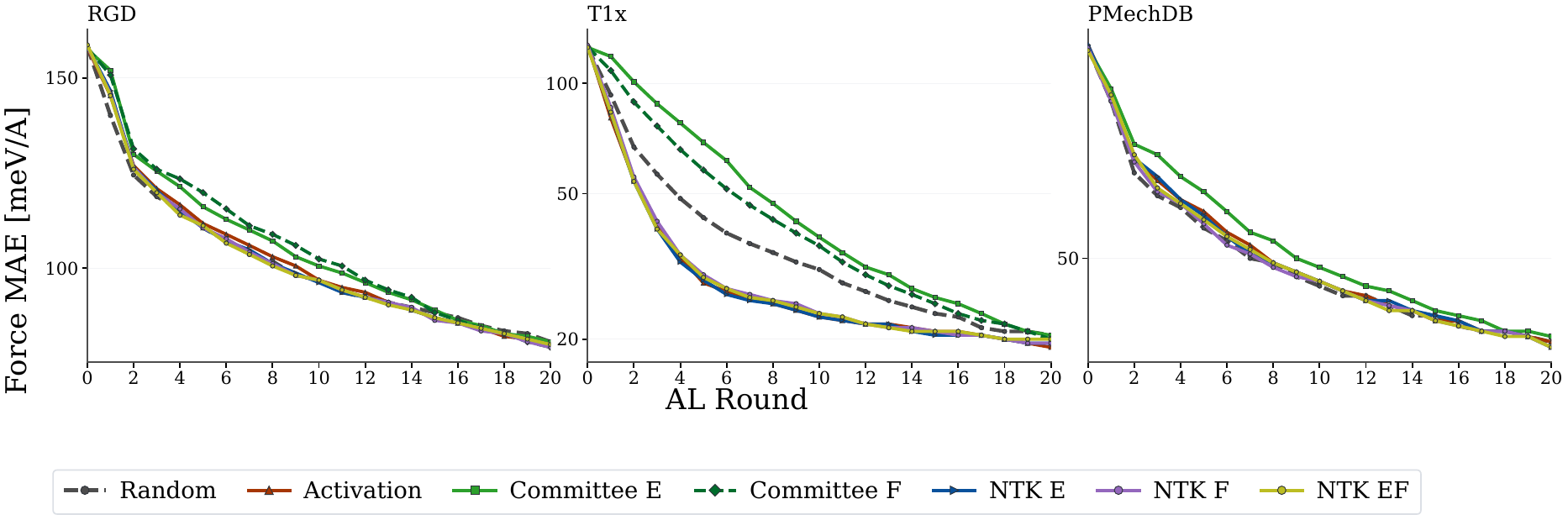}
  \caption{Force MAE learning curves (meV/\AA) for the methods presented in the main text.}
  \label{fig:mae_f}
\end{figure}

\clearpage

%\subsection{Energy-force weighting}
%\input{version1_tables/efweighting_table}
%\clearpage

\section{Details of controlled biasing}
\label{sec:appendix-t1x-case-study}

T1x provides a controlled setting for studying offline active learning under
candidate-pool bias. Each reaction family contains structures along NEB and
climbing-image NEB paths, so the frame index gives a discrete proxy for
position along the reaction coordinate. This lets us separate two practically
important forms of pool shift: imbalance across reaction families and imbalance
within a reaction pathway.

Formally, for reaction family $r$, let
$\mathcal{X}_r = \{x_{r,i,f}\}$, where $i$ indexes reaction instances or
optimization trajectories and $f \in \{1, \dots, F_{r,i}\}$ indexes frames
along the path. Each point $x_{r,i,f}$ is a molecular geometry with associated
energy and force labels. Let $N_r = |\mathcal{X}_r|$ denote the number of
available structures for reaction family $r$. Uniform sampling over all
structures induces $p(r) \propto N_r$, producing the empirical reaction-family
imbalance that we refer to as \textit{natural bias}.

To construct controlled \textit{inter-reaction bias}, we introduce
reaction-level weights $\mathbf{w}=(w_1,\ldots,w_R)$ and sample reaction
families according to $p(r) \propto w_r N_r$, while sampling frames uniformly
within each chosen reaction. To construct controlled \textit{intra-reaction
bias}, we fix a reaction family and alter the frame distribution within that
family. For a trajectory of length $F_{r,i}$, frame weights
$\{\pi_f\}_{f=1}^{F_{r,i}}$ define
\[
    p(f \mid r,i) \propto \pi_f,
    \quad f \in \{1,\dots,F_{r,i}\}.
\]
In all settings, the test set is fixed and balanced, so differences in
performance reflect robustness to the acquisition pool rather than changes in
the evaluation distribution.

All biasing experiments use the same offline active-learning protocol. We randomly sample 5 pathways from T1x, giving approximately 4.2k structures, and fix a balanced test/validation set of 175/50 structures. For inter-reaction bias, we up-weight individual reaction families by a factor of 5 relative to a uniform (1:1:1:1:1) baseline. For intra-reaction bias, we similarly upweight specific frames (0th, 5th, and 8th) by a factor of 5 relative to uniform sampling along the reaction coordinate. We perform 40 active-learning rounds, acquiring 5 points per round, starting from an initial seed set of 5 structures, and finetuning a SPICE-2-pretrained MACE model after each round.

\subsection{Inter-reaction bias}

\begin{figure}[h]
  \centering
  \includegraphics[width=\linewidth]{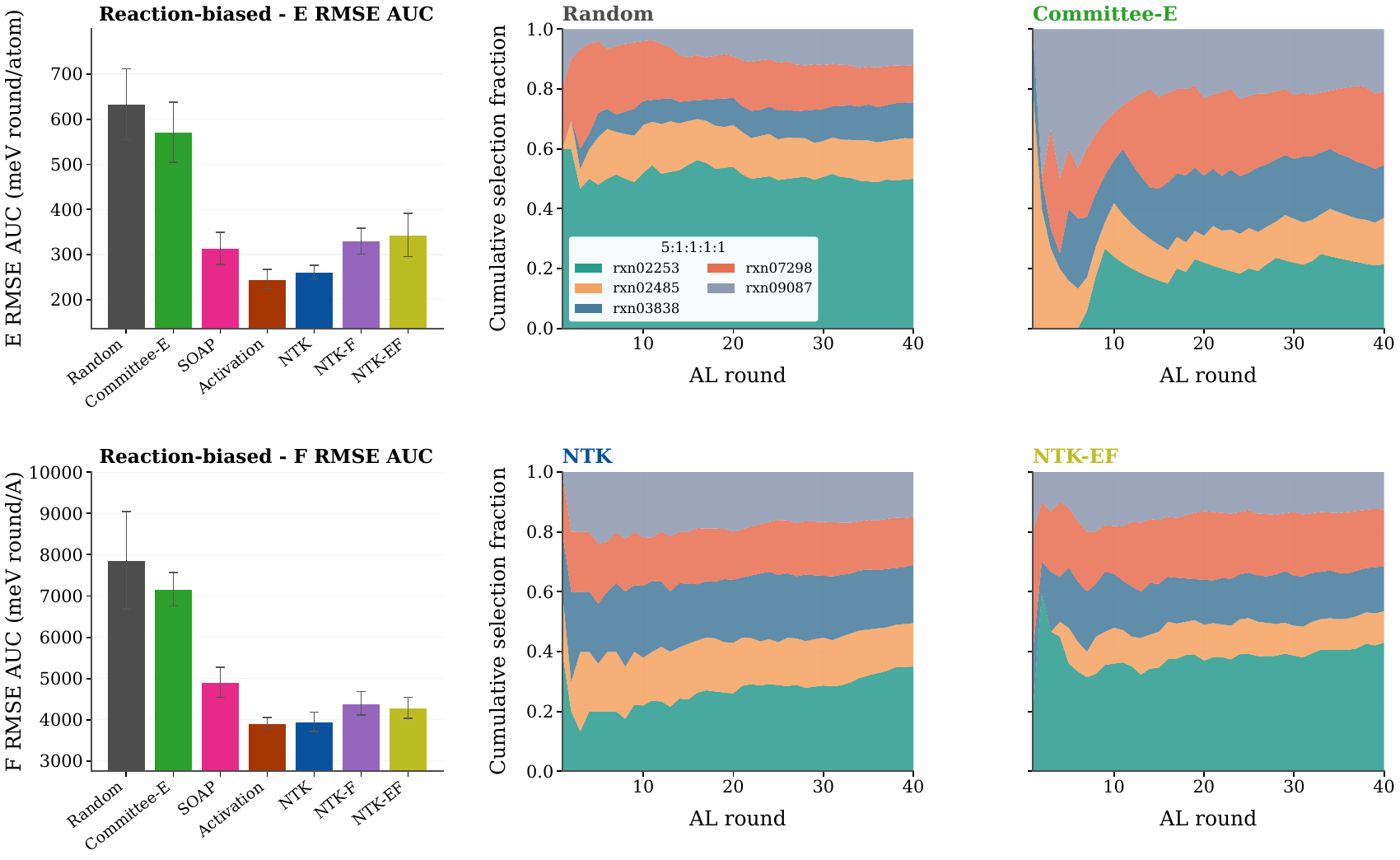}
  \caption{Inter-reaction bias. Left: energy RMSE AUC (meV$\cdot$round/atom) and force RMSE AUC (meV$\cdot$round/\AA) under reaction-weighted candidate pools. Error bars denote the standard deviation across different biasing configurations. Right: cumulative selection fraction across reaction families over active learning rounds.}
  \label{fig:inter-reaction-bias-summary}
\end{figure}

Figure~\ref{fig:inter-reaction-bias-summary} summarizes the inter-reaction bias results. The left
panels report energy and force RMSE AUC, while the right panels show the
cumulative selection fraction across reaction families. Across both metrics,
model-based methods substantially improve over random selection, with
Activation and NTK-based approaches achieving the lowest AUC. Force-aware
variants (NTK-F, NTK-EF) remain competitive, indicating that incorporating
force information does not degrade performance under this form of bias.

The selection distributions highlight differences in behaviour. Random
selection largely reflects the imposed reaction imbalance, while Committee-E
shows unstable allocation across reaction families. In contrast, NTK and
NTK-EF distribute selections more evenly, indicating reduced sensitivity to
the skewed pool. Overall, learned representation-based methods maintain more
stable behaviour under inter-reaction bias than committee-based disagreement.

\clearpage

\subsection{Intra-reaction bias}

The intra-reaction setting is a finer-grained test of robustness. Instead of
changing the reaction-family distribution, we fix the reaction family and skew
the candidate pool along the reaction coordinate by changing the frame weights
$\pi_f$. This makes the benchmark harder: structures within a single reaction
manifold can be highly similar, so acquisition must resolve subtle geometric
differences rather than simply cover distinct reaction classes. In practice, we
bias frames 1--8; frames 9 and 10 are excluded from the biasing sweep because
they typically contain fewer than 10 structures, making them too sparse for a
stable controlled-bias comparison.

\begin{figure}[h]
  \centering
  \includegraphics[width=\linewidth]{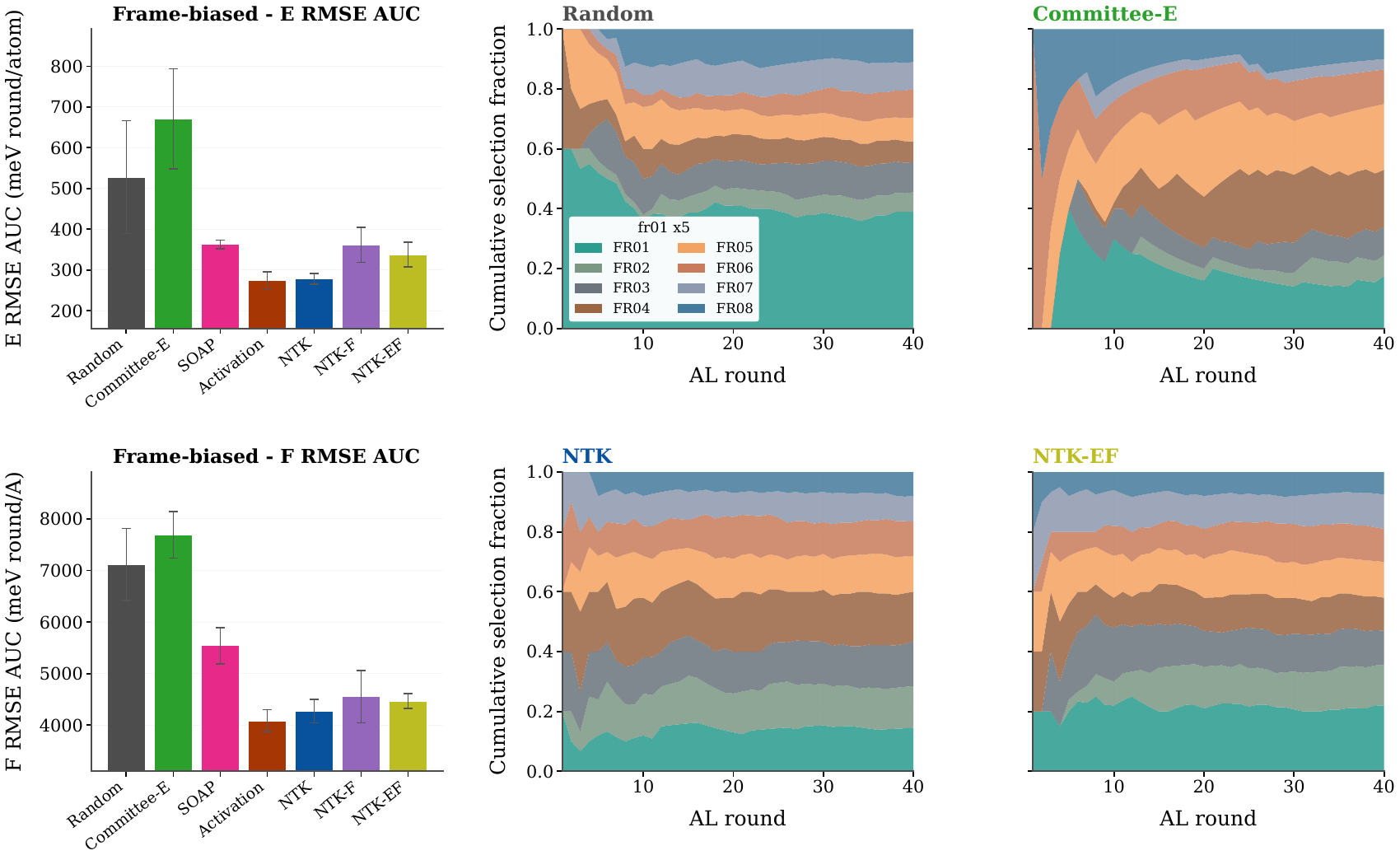}
  \caption{Intra-reaction (frame) bias. Left: energy RMSE AUC (meV$\cdot$round/atom) and force RMSE AUC (meV$\cdot$round/\AA) under frame-skewed candidate pools. Error bars denote the standard deviation across different biasing configurations. Right: cumulative selection fraction across frames over active learning rounds.}
  \label{fig:intra-reaction-bias-summary}
\end{figure}

Figure~\ref{fig:intra-reaction-bias-summary} summarizes performance under
frame-biased pools. The left panels report energy and force RMSE AUC, while the
right panels show the cumulative selection fraction across reaction coordinates.
Across both metrics, model-based methods improve substantially over random
selection, with Activation and NTK-based approaches achieving the lowest AUC.
Force-aware variants (NTK-F, NTK-EF) remain competitive, indicating that
incorporating force information does not degrade performance in this setting.

The selection distributions highlight differences in behaviour. Random selection
tracks the imposed bias, while Committee-E exhibits unstable allocation across
frames. In contrast, NTK and NTK-EF produce more balanced selections along the
reaction coordinate, indicating that they are less sensitive to the skewed
candidate distribution. Overall, this setting shows that when variation is
confined to a single reaction manifold, learned representation-based methods
provide more stable behaviour than committee-based disagreement.

\clearpage

Table~\ref{tab:bias_summary} reports the average AUC on the test set and the maximum deviation across bias settings. Across both regimes, model-based methods outperform random selection, with Activation and NTK-based approaches achieving the lowest AUC, and force-aware variants (NTK-F, NTK-EF) remaining competitive. Descriptor-based methods provide moderate improvements but are generally less competitive than learned representations, while Committee-E shows higher AUC and larger deviations across bias settings, indicating greater sensitivity to candidate-pool shift. 
\begin{table}[h]
\centering
\scriptsize
\setlength{\tabcolsep}{8pt}
\caption{Controlled-bias summary metrics, reported as mean [max deviation]
across bias settings. Final columns are final-round RMSE in meV for energy and
meV\,\AA$^{-1}$ for forces. AUC is the discrete sum of RMSE over acquisition
steps. Best (lowest) values are in bold. All descriptor based methods use LCMD acquisition.}
\label{tab:bias_summary}
\begin{tabular}{llcccc}
\toprule
Bias & Method & Energy AUC ($\downarrow$) & Force AUC ($\downarrow$) & Final E RMSE ($\downarrow$) & Final F RMSE ($\downarrow$)\\
\midrule
Inter (Reaction)
& Random      & 633.3 [129.7] & 7865.2 [1807.8] & 6.3 [3.7] & 123.7 [65.3] \\
& Committee-E & 571.7 [108.7] & 7165.5 [543.5]  & 4.5 [2.5] & 83.8 [12.8] \\
& Tanimoto    & 587.2 [168.8] & 6973.8 [1498.2] & 8.5 [3.5] & 118.3 [25.7] \\
& SOAP        & 313.2 [72.8]  & 4909.8 [816.2]  & \textbf{2.8 [0.8]} & 58.3 [11.3] \\
& Activation  & \textbf{245.3 [35.3]} & \textbf{3916.8 [217.8]} & 3.2 [0.8] & 56.0 [11.0] \\
& NTK         & 261.3 [22.3]  & 3945.3 [393.3]  & \textbf{2.8 [1.2]} & \textbf{54.3 [11.3]} \\
& NTK-F       & 329.7 [41.3]  & 4395.5 [467.5]  & 3.5 [0.5] & 59.5 [13.5] \\
& NTK-EF      & 343.2 [71.2]  & 4287.7 [352.3]  & 3.3 [1.3] & 59.7 [13.7] \\
\addlinespace[2pt]
\hline
\hline
\addlinespace[2pt]
Intra (Frame)
& Random      & 528.3 [164.8] & 7117.8 [1018.3] & 6.5 [2.5] & 112.0 [19.0] \\
& Committee-E & 671.0 [210.0] & 7690.5 [722.5]  & 5.0 [2.0] & 90.5 [12.5] \\
& Tanimoto    & 492.0 [139.0] & 6255.0 [623.0]  & 7.0 [3.0] & 112.5 [26.5] \\
& SOAP        & 362.8 [14.8]  & 5541.3 [388.8]  & 3.5 [1.5] & 67.5 [6.5] \\
& Activation  & \textbf{274.5 [30.5]} & \textbf{4086.5 [318.5]} & 3.5 [0.5] & 60.5 [10.5] \\
& NTK         & 279.0 [22.0]  & 4269.8 [261.3]  & 4.0 [1.0] & 62.3 [13.8] \\
& NTK-F       & 361.8 [56.3]  & 4551.0 [578.0]  & \textbf{3.0 [0.0]} & \textbf{59.0 [12.0]} \\
& NTK-EF      & 338.0 [41.0]  & 4466.3 [196.3]  & 3.3 [0.8] & 60.0 [10.0] \\
\bottomrule
\end{tabular}
\end{table}

\clearpage

\subsection{NTK force kernel plots}\label{app:force_ntk_plot}
\begin{figure*}[h]
    \centering
    \begin{subfigure}[h]{0.32\textwidth}
      \centering
      \includegraphics[width=\linewidth]{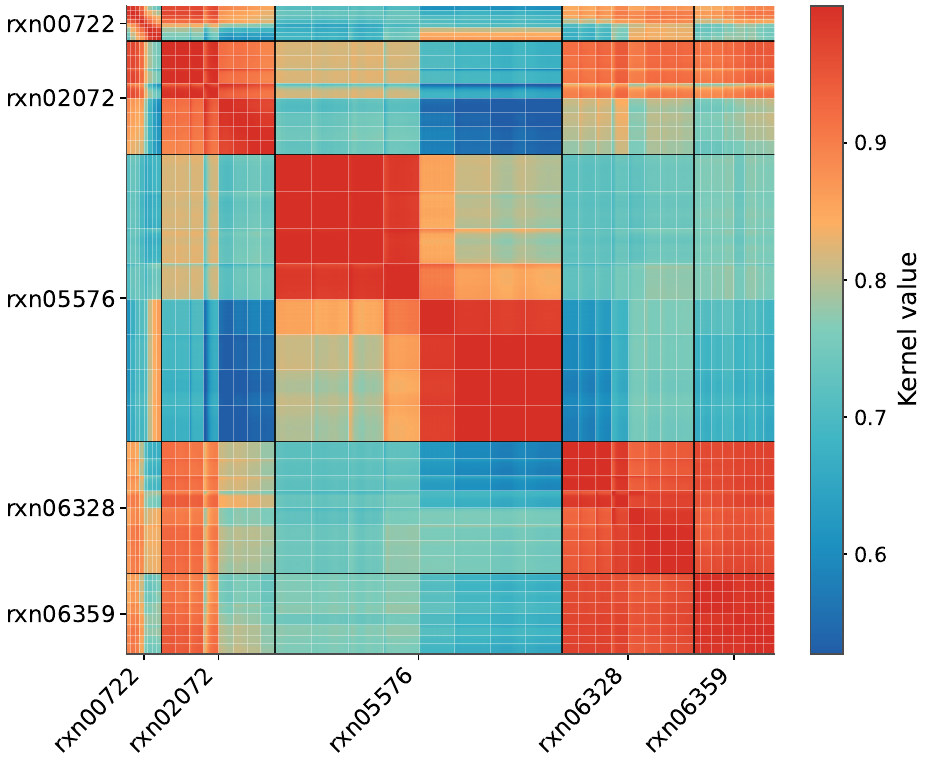}
      \caption{Energy NTK}
    \end{subfigure}
    \hfill
    \begin{subfigure}[h]{0.32\textwidth}
      \centering
      \includegraphics[width=\linewidth]{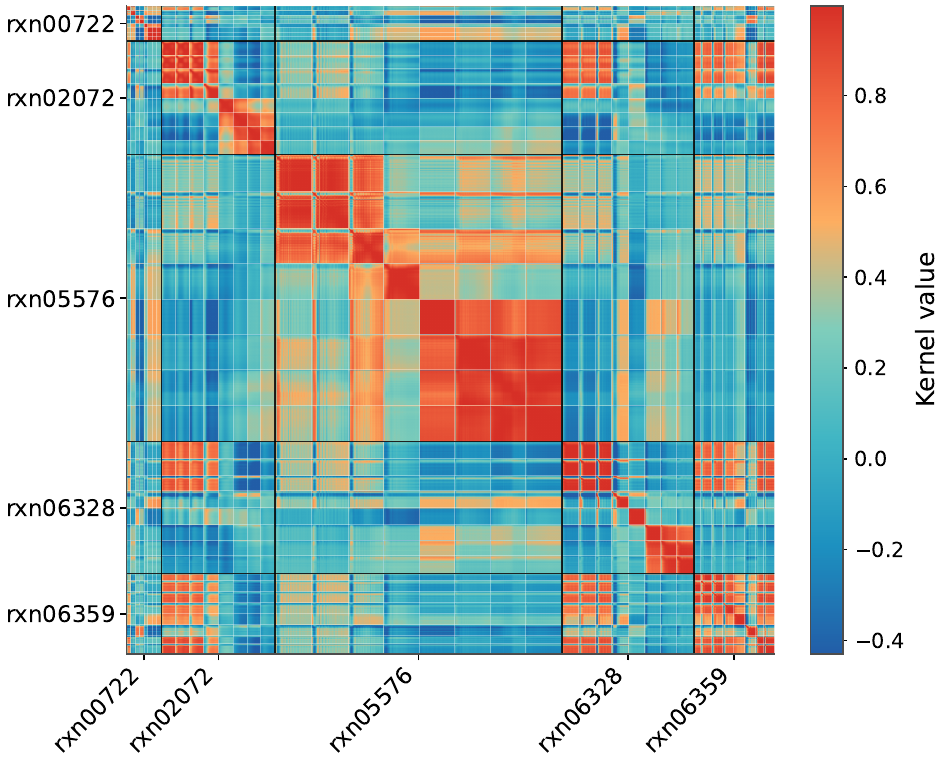}
      \caption{Force NTK}
    \end{subfigure}
    \hfill
    \begin{subfigure}[h]{0.32\textwidth}
      \centering
      \includegraphics[width=\linewidth]{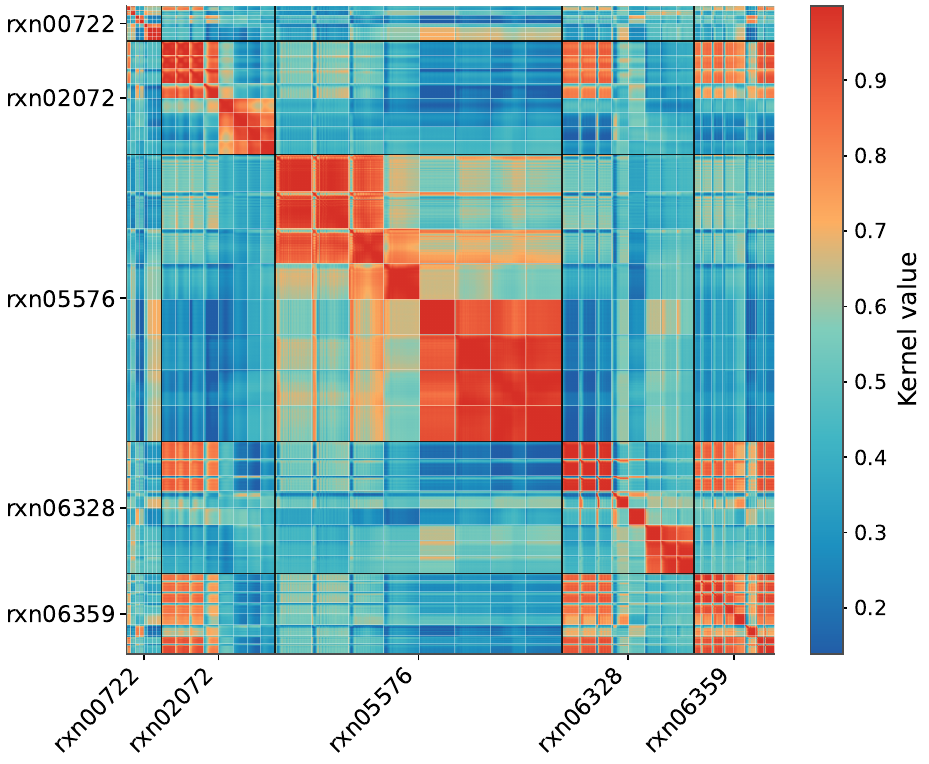}
      \caption{Energy-Force NTK}
    \end{subfigure}
    \caption{Global kernel matrices on a five-reaction T1x subset. Structures are
    sorted by reaction family and frame index. The Energy-Force NTK is an equal weighting ($w_E=w_F=0.5$).}
    \label{fig:force-kernel-geometry}
  \end{figure*}

In this section, we visualise the structure of the force-aware NTK by comparing it to the energy NTK on 5 randomly selected reactions from T1x. In Figure~\ref{fig:force-kernel-geometry}, the kernel matrix is ordered first by reaction ID and then by frame index along each reaction pathway. Both kernels exhibit structured block patterns; however, the force-aware kernel more clearly separates regions corresponding to reactant, transition-state, and product configurations, as reflected in the finer structure within each block.

Figure \ref{fig:force-kernel-geometry} hints at the importance of how energy and force representations are combined in joint kernels. In particular, the combined kernels visually resemble the force NTK more strongly than the energy NTK, indicating that the force contribution may dominate the joint representation despite cosine normalization. This suggests that the relative weighting and normalization of energy and force features are important design choices. More sophisticated magnitude-alignment strategies such as standard normalization prior to kernel combination could thus potentially yield better-balanced joint representations and improve uncertainty estimation performance.

\section{Additional OC20 information}\label{app:oc20}

This appendix collects experimental details and additional results for the OC20 benchmark of Section~\ref{sec:oc20}.

\subsection{Learning curves}\label{app:oc20-learning-curves}

\begin{figure}[h]
  \centering
  \includegraphics[width=\linewidth]{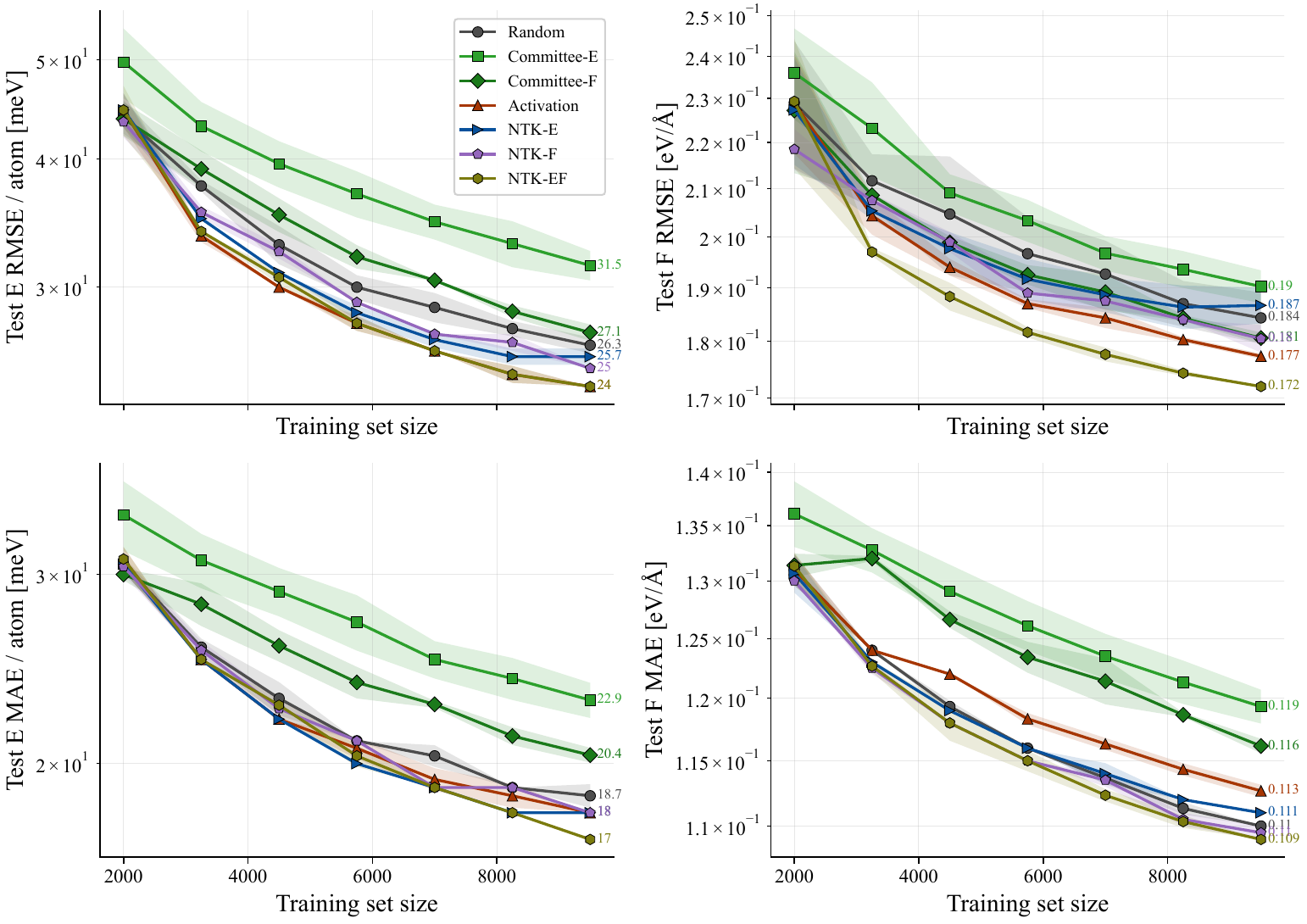}
  \caption{OC20 test error vs.\ training-set size on the $20{,}000$-structure aggregated test set. Columns: energy and force RMSE/MAE (energy reported per atom, in meV; forces in eV/\AA). Shaded bands are across-seed RMS deviations; the value at the right margin annotates each method's final-round error.}
  \label{fig:oc20_learning_curves}
\end{figure}

Figure~\ref{fig:oc20_learning_curves} shows the AL learning curves on the $20{,}000$-structure test set for the headline method subset (Random, Committee-E, Committee-F, Activation, NTK-E, NTK-F, NTK-EF), as a function of training-set size from the initial $2{,}000$ to the final $9{,}500$ labelled structures. The joint NTK-EF achieves the lowest test error from the first acquisition round onward across all four metrics (energy and force RMSE/MAE), and the gap relative to the energy-only and force-only NTK variants persists across the full schedule. Both committee variants stay above all NTK methods on every metric, with Committee-E showing the largest deviation on energy errors. Activation tracks the NTK family closely on energy metrics but trails on force errors.

\subsection{Per-split test errors at the final round}\label{app:oc20-split-metrics}

\begin{figure}[h]
  \centering
  \includegraphics[width=\linewidth]{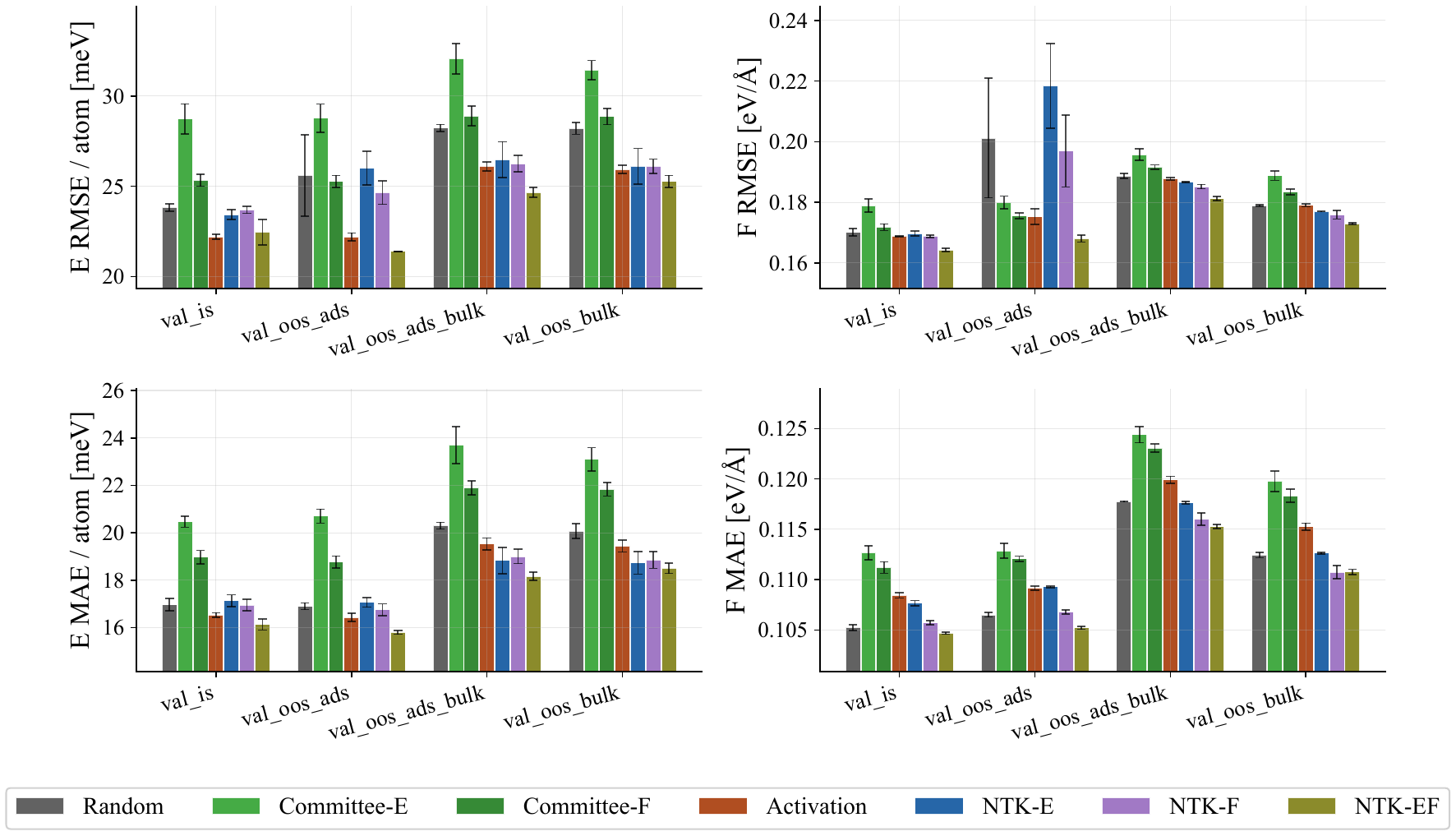}
  \caption{OC20 final-round test errors resolved by validation split (\texttt{val\_is}, \texttt{val\_oos\_ads}, \texttt{val\_oos\_ads\_bulk}, \texttt{val\_oos\_bulk}). Bars are across-seed means; error bars are across-seed RMS deviations. Y-axes are zoomed to the data range so that method differences remain visible.}
  \label{fig:oc20_split_metrics}
\end{figure}

Figure~\ref{fig:oc20_split_metrics} resolves the final-round test errors of Figure~\ref{fig:oc20_finalbars} into all four OC20 splits, rather than aggregating the three OOD splits into a single OOD column. The ranking of methods is consistent across splits: NTK-EF is best on every (metric, split) pair on energy errors and ties with NTK-F within seed noise on the force splits. The OOD splits amplify the spread between methods most strongly on the energy axis: Committee-E rises from $\approx\!29$ to $\approx\!32$\,meV/atom on E~RMSE between \texttt{val\_is} and \texttt{val\_oos\_ads\_bulk}, while NTK-EF stays close to $25$\,meV/atom across the four splits. The energy-only NTK degrades sharply on \texttt{val\_oos\_ads}, consistent with the discussion in Section~\ref{sec:oc20} that energy features alone do not distinguish adsorbate chemistry. %and the diversity objective then emphasises bulk-surface geometry rather than adsorbate variation.

\clearpage
\section{Computational considerations: runtime and memory}
\label{app:computational-considerations}

This appendix contains per-method runtime and memory measurements behind Figure~\ref{fig:pareto_combined}, and parameter-subset analysis referenced in Section~\ref{sec:scaling}.

%Numbers are extracted from active-learning run logs and reported as steady-state averages over AL rounds 2--5. We do this since for our logs round 0 is initial training, round 1 is JIT-cold, and round 6 has no acquisition pass since it evaluates the final model. Two scales are covered: the small T1x pools ($n_{\mathcal{P}} \sim 4$k), where the full candidate--candidate kernel is comfortably materialisable, and the OC20 pool ($n_{\mathcal{P}} \approx 197$k), where it is not.

\subsection{OC20 per-method timings}\label{app:timing-oc20}
Table~\ref{tab:oc20_timings} reports acquisition wall-clock, training wall-clock, and peak memory usage on the OC20 candidate pool (\(n_{\mathcal{P}} \approx 200\)k). At this scale, direct candidate--candidate kernel materialisation is infeasible, so all NTK and activation-based methods use the chunked shortlist-based PV pipeline described in Section~\ref{sec:acquisition}. Timings are reported as steady-state averages over active-learning rounds, which starts from round 2, since the initial round is initial training, and round 1 is jit-cold \cite{jax2018github}.

%The OC20 candidate pool ($n_{\mathcal{P}} \approx 197{,}000$, NTK feature dimension $d_{\mathrm{NTK}} = Z_{\max} \cdot c = 88 \times 128 = 11{,}264$ for the embeddings subset, $d_{\mathrm{Act}} = 256$ for the activation embedding) is too large for direct kernel materialisation: the float32 candidate--candidate kernel alone would require $\approx\!145$\,GB. All NTK and Activation methods therefore use the chunked shortlist-then-select pipeline of Algorithm~\ref{alg:chunked_pv} with chunk size $c = 5{,}000$, shortlist $K = 50{,}000$, and acquisition batch $1{,}250$. Table~\ref{tab:oc20_timings} reports per-round acquisition wall-clock, surrogate retraining wall-clock, and peak GPU memory.

\begin{table}[h]
\centering
\small
\setlength{\tabcolsep}{8pt}
\caption{OC20 per-round acquisition and training cost (mean \(\pm\) std across seeds). Timings are reported as steady-state averages over active-learning rounds 2-5, excluding the initial training round and the jit-cold first acquisition round. ``Acq.'' is the time spent inside the acquisition function itself.}
\label{tab:oc20_timings}
\begin{tabular}{lrrr}
\toprule
Method & Acq. (s) & Train (s) & Total (s) \\
\midrule
NTK-EF (embeddings) LCMD       & $8317 \pm 171$  & $1924 \pm 27$   & $10242 \pm 170$  \\
NTK-EF (readouts) LCMD     & $3845 \pm 15$   & $1840 \pm 11$   & $5685 \pm 4$     \\
NTK-F (embeddings) LCMD   & $5890 \pm 175$  & $1729 \pm 27$   & $7619 \pm 148$   \\
NTK-F (readouts) LCMD          & $3025 \pm 107$  & $1850 \pm 148$  & $4874 \pm 40$    \\
NTK-E (embeddings) LCMD        & $5188 \pm 1541$ & $1736 \pm 79$   & $6925 \pm 1466$  \\
NTK-E (embeddings) PV          & $4168 \pm 1456$ & $1829 \pm 5$    & $5997 \pm 1456$  \\
NTK-E (readouts) LCMD      & $875 \pm 30$    & $1843 \pm 108$  & $2718 \pm 77$    \\
Activation LCMD                & $613 \pm 34$    & $1804 \pm 150$  & $2417 \pm 164$   \\
Activation PV                  & $982 \pm 316$   & $1956 \pm 104$  & $2937 \pm 276$   \\
Committee-E ($M{=}3$)          & $4707^{\dagger} \pm 474$  & $8890 \pm 947$ & $13596 \pm 1420$ \\
Committee-F$^{\ast}$ ($M{=}3$) & --              & --              & $13119 \pm 496$  \\
Random                         & $<1$            & $2323 \pm 134$  & $2323 \pm 134$   \\
\bottomrule
\end{tabular}
\begin{flushleft}
\footnotesize $^{\dagger}$For committee acquisition, each member rebuilds the candidate-pool graphs independently, and re-JITs the force field, and the $M{=}3$ forward passes run sequentially. Graph caching and \texttt{vmap}-fused implementation could reduce this fairly significantly, and could push the acquisition time below that of NTK methods, but the training (and overall wall-time) we expect to remain larger even in a more efficient implementation.\\
$^{\ast}$Committee-F shares Committee-E's timings: the MACE forward pass returns energy and forces jointly, so the only difference between the two is whether the per-structure score aggregates energy disagreement or per-atom force disagreement; the cost of acquisition and training is identical.
\end{flushleft}
\end{table}

We report both PV and LCMD-based acquisition. PV selects candidates directly according to their feature-space posterior variance, while LCMD first constructs a high-uncertainty shortlist using PV and then applies diversity-based selection on the shortlist. All methods operate on the same candidate pool using the chunked feature-space pipeline; the primary difference between activation and NTK-based methods is therefore the representation dimension (\(d_{\mathrm{Act}} = 256\) versus \(d_{\mathrm{NTK}} = 11{,}264\) for the embeddings subset).

Both committees train $M{=}3$ ensemble members plus a 4th evaluation model and run the same forward passes on the pool, differing only in whether energy or force disagreement is aggregated. Since the evaluation model is not necessary, one could further divide the train times by a factor $4/3$ without changing the qualitative results. Committee is still Pareto dominated on both fronts. When the experiments were launched, committee-F acquisition and train times were not logged, so the only time shown in Table \ref{tab:oc20_timings} is the total time..

\subsection{Memory footprint and scaling}\label{app:memory}
Table~\ref{tab:memory_pareto} shows the breakdown of peak memory usage for the chunked vs.\ non-chunked PV implementation across pool sizes.
\begin{table}[h]
\centering
\small
\setlength{\tabcolsep}{8pt}
\caption{Acquisition peak memory overhead relative to the pre-acquisition baseline for chunked and non-chunked feature-space PV at varying pool sizes \(n_S = n_P + n_T\), using the OC20 NTK-E (embeddings) representation. The non-chunked implementation materialises the full \(n_S \times n_S\) kernel and exhausts host RAM beyond \(n_S \approx 6 \times 10^{4}\), whereas the chunked implementation scales across the full tested range. ``--'' indicates configurations that were not measured, and ``OOM (host)'' indicates that the implementation exhausted host RAM before completing.}
\label{tab:memory_pareto}
\begin{tabular}{rrr}
\toprule
$n_S$ & chunked PV (GiB) & non-chunked PV (GiB) \\
\midrule
$500$       & $0.35$ & $0.41$ \\
$1{,}000$   & $0.48$ & $0.63$ \\
$5{,}000$   & $2.72$ & $2.41$ \\
$10{,}000$  & $3.29$ & $4.92$ \\
$25{,}000$  & --     & $14.56$ \\
$40{,}000$  & --     & $26.11$ \\
$50{,}000$  & $4.09$ & $35.33$ \\
$60{,}000$  & --     & $44.63$ \\
$200{,}000$ & $5.64$ & OOM (host) \\
\bottomrule
\end{tabular}
\end{table}

\subsection{Parameter-subset trade-off on T1x}\label{app:timing}

Table~\ref{tab:t1x_weight_subsets} compares NTK parameter subsets on a 5-reaction T1x subset using full LCMD acquisition. We report energy and force RMSE together with per-pass NTK runtime.

\begin{table}[h]
\centering
\small
\setlength{\tabcolsep}{8pt}
\caption{NTK parameter-subset cost--accuracy trade-offs on T1x. Each variant restricts the NTK to gradients with respect to a single MACE parameter block of the SPICE-2 backbone, using full (non-shortlisted) LCMD acquisition on a balanced 5-reaction T1x subset (\(4{,}196\) candidates). Per-pass NTK time reports the post-JIT runtime, excluding one-time JAX compilation overhead. Test errors are averaged over the final 5 active-learning rounds, with \(\pm\) denoting the across-seed standard deviation. Energy errors are reported in meV/atom and force errors in eV/\AA. Lowest values in each column are highlighted in bold.}
\label{tab:t1x_weight_subsets}
\begin{tabular}{lrrcc}
\toprule
Parameter subset & Parameters & NTK pass [s] & RMSE E/atom [meV] & RMSE F [eV/\AA{}] \\
\midrule
readouts        & 2{,}192      & \textbf{1.07 $\pm$ 0.06} & 2.6 $\pm$ 0.4 & \textbf{0.0432 $\pm$ 0.0004} \\
embeddings      & 1{,}920      & 3.97 $\pm$ 0.12          & \textbf{2.4 $\pm$ 0.3} & 0.0462 $\pm$ 0.0020 \\
interactions    & 762{,}880    & 11.93 $\pm$ 0.64         & \textbf{2.4 $\pm$ 0.3} & 0.0445 $\pm$ 0.0032 \\
last\_layer     & 1{,}372{,}160 & 12.07 $\pm$ 0.91         & \textbf{2.4 $\pm$ 0.0} & 0.0442 $\pm$ 0.0026 \\
\bottomrule
\end{tabular}
\end{table}

Runtime increases substantially with feature dimension, ranging from \(\sim\!1\)\,s for readouts to \(\sim\!12\)\,s for the largest parameter subsets. Despite this, all subsets achieve comparable energy accuracy within seed noise, while the lightweight readouts and embeddings subsets remain competitive on force accuracy. On this small study, larger subsets therefore provide little measurable accuracy benefit relative to their computational cost.

In practice, chunked feature-space PV requires storing a \(d \times d\) posterior-covariance matrix, making feature dimension the dominant scalability bottleneck. As a result, only lightweight parameter subsets such as embeddings and readouts remain practical at OC20 scale, while larger subsets would require additional dimensionality reduction. For reference, a float32 \(d \times d\) matrix with \(d = 10^5\) requires approximately \(4 \times 10^{10}\) bytes of memory (\(\sim\!40\)\,GB or \(37.3\)\,GiB), meaning that the posterior-covariance matrix alone would occupy roughly half of an \(80\)\,GB GPU.

\section{Broader Impact}\label{app:broader_impact}
This work develops scalable, force-aware active learning methods for machine-learning interatomic potentials (MLIPs), aimed at reducing the cost of curating training data for atomistic simulation. The positive societal impacts are primarily indirect: more data-efficient MLIPs lower the computational and financial cost of accurate atomistic modelling, which underpins applications such as catalyst discovery for clean energy and carbon capture, battery and electrolyte design, drug discovery, and the development of novel functional materials. By reducing the number of expensive DFT calculations required to fine-tune foundation models for new chemical domains, our methods can also lower the energy footprint of large-scale atomistic dataset construction and broaden access to high-quality MLIPs for research groups without large compute budgets. We do not foresee direct negative societal impacts specific to this work, as it concerns a methodological improvement to data selection given already generated structures rather than the introduction of new capabilities or models.

\end{document}